\documentclass[0.5pt,letterpaper]{article}
\usepackage{fullpage}

\usepackage{amsmath,amsfonts,bm}

\def\eqref#1{equation~\ref{#1}}
\def\1{\bm{1}}

\DeclareMathAlphabet{\mathsfit}{\encodingdefault}{\sfdefault}{m}{sl}
\SetMathAlphabet{\mathsfit}{bold}{\encodingdefault}{\sfdefault}{bx}{n}

\newcommand{\ldim}{d}
\newcommand{\hdim}{m}
\newcommand{\RR}{\mathbb{R}}
\newcommand{\w}{\omega}
\newcommand{\x}{\bm{x}}
\newcommand{\X}{\bm{X}}
\newcommand{\y}{\bm{y}}
\newcommand{\Y}{\bm{Y}}
\newcommand{\err}{\mathcal{E}}
\newcommand{\A}{\bm{A}}
\newcommand{\GS}{\text{GS}}
\newcommand{\loss}{\ell}
\newcommand{\T}{\mathcal{T}}

\usepackage{hyperref}
\usepackage{url}
\usepackage{times}
\usepackage{helvet}
\usepackage{courier}
\usepackage{amsmath, amsfonts, amssymb}
\usepackage{mathtools}
\usepackage{subfigure}
\usepackage[group-separator={,}]{siunitx}
\usepackage{booktabs}
\usepackage{color}
\usepackage{float}
\usepackage{wrapfig}
\usepackage{lipsum}
\usepackage{tabularx}
\usepackage{array}
\usepackage{natbib}

\usepackage[multiple]{footmisc}
\newcommand{\Blue} [1]{{\color{blue} {#1}}}
\usepackage[normalem]{ulem}
\newcommand{\stkout}[1]{\ifmmode\text{\sout{\ensuremath{#1}}}\else\sout{#1}\fi}

\title{TriMap: Large-scale Dimensionality Reduction Using Triplets}

\author{Ehsan Amid \& Manfred K. Warmuth\\
Google Research, Brain Team\\
\texttt{\{eamid, manfred\}@google.com}
}
\date{}

\begin{document}

\maketitle

\begin{abstract}
We introduce ``TriMap''; a dimensionality reduction technique based on triplet constraints, which preserves the global structure of the data better than the other commonly used methods such as t-SNE, LargeVis, and UMAP. To quantify the global accuracy of the embedding, we introduce a score that roughly reflects the relative placement of the clusters rather than the individual points. We empirically show the excellent performance of TriMap on a large variety of datasets in terms of the quality of the embedding as well as the runtime. On our performance benchmarks, TriMap easily scales to millions of points without depleting the memory and clearly outperforms t-SNE, LargeVis, and UMAP in terms of runtime.

[Update] The results in the current version of the paper is using an older version of the code available at:~\url{https://github.com/eamid}. The results will be updated using version $\geq$ 1.1.0 (although there will not be substantial changes in terms of quality). A JAX implementation of TriMap is also available at:~\url{https://github.com/google/trimap}.
\end{abstract}

\section{Introduction}

Data visualization based on dimensionality reduction (DR) is a core problem in data analysis and machine learning.
The aim of DR is to provide a low-dimensional representation
(typically in 2D or 3D) of a given high-dimensional dataset
that preserves the overall structure of the data as much as possible.
The earlier approaches for DR involve linear methods such as PCA~\citep{pca}.
PCA aims to maintain the second-order statistics of the data by projecting the points into the low dimensional space that preserves the maximum amount of variance among all such projections.
As a result, PCA has been shown to be effective in preserving the \emph{global structure} of the data~\citep{global}. The global structure includes the overall shape of the dataset, placement of the clusters, and the existence of potential outliers.
Unlike PCA, much of the focus of the more recent non-linear methods, including t-SNE~\citep{tsne}, LargeVis~\citep{lv}, and UMAP~\citep{umap} has been on preserving the local neighborhood structure of each individual point. Similarly, the common performance measures of DR such as trustworthiness-continuity~\citep{tc}, precision-recall (i.e., AUC)~\citep{auc}, and nearest-neighbor accuracy have also been developed by retaining the same focus on reflecting the local accuracy of the embedding. Thus, there has been a lack of attention on developing methods that focus on preserving the global structure of the data and, likewise, practical performance measures to assess the global accuracy.

\begin{table}[h!]
\vspace{-1.2cm}
\setlength{\tabcolsep}{1pt}
\renewcommand{\tablename}{Figure}
\begin{center}
\resizebox{0.9\textwidth}{!}{
\begin{tabular}{cc}
     {\footnotesize FC1 - t-SNE \texttt{(NN\,=\,$\bm{0.87}$,GS\,=\,$0.58$)}} & {\footnotesize FC1 - TriMap \texttt{(NN\,=\,$0.72$,GS\,=\,$\bm{0.70}$)}}\\[-1mm]
    \subfigure{\includegraphics[ width=0.45\textwidth]{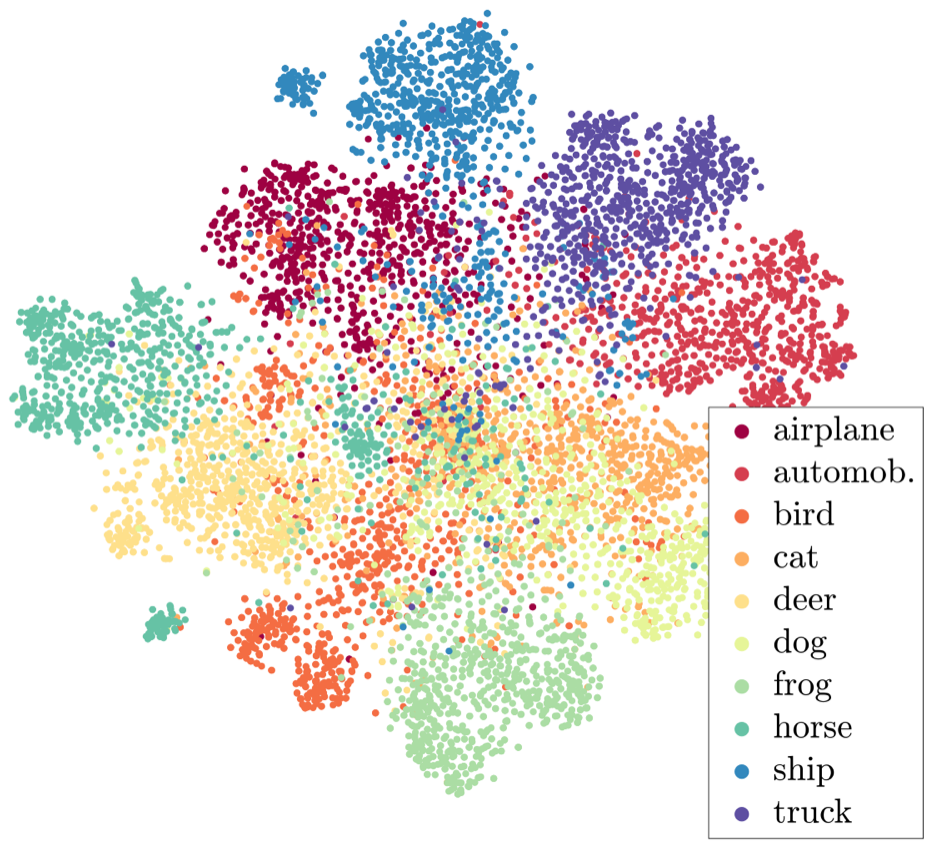}} &
    \subfigure{\includegraphics[ width=0.45\textwidth]{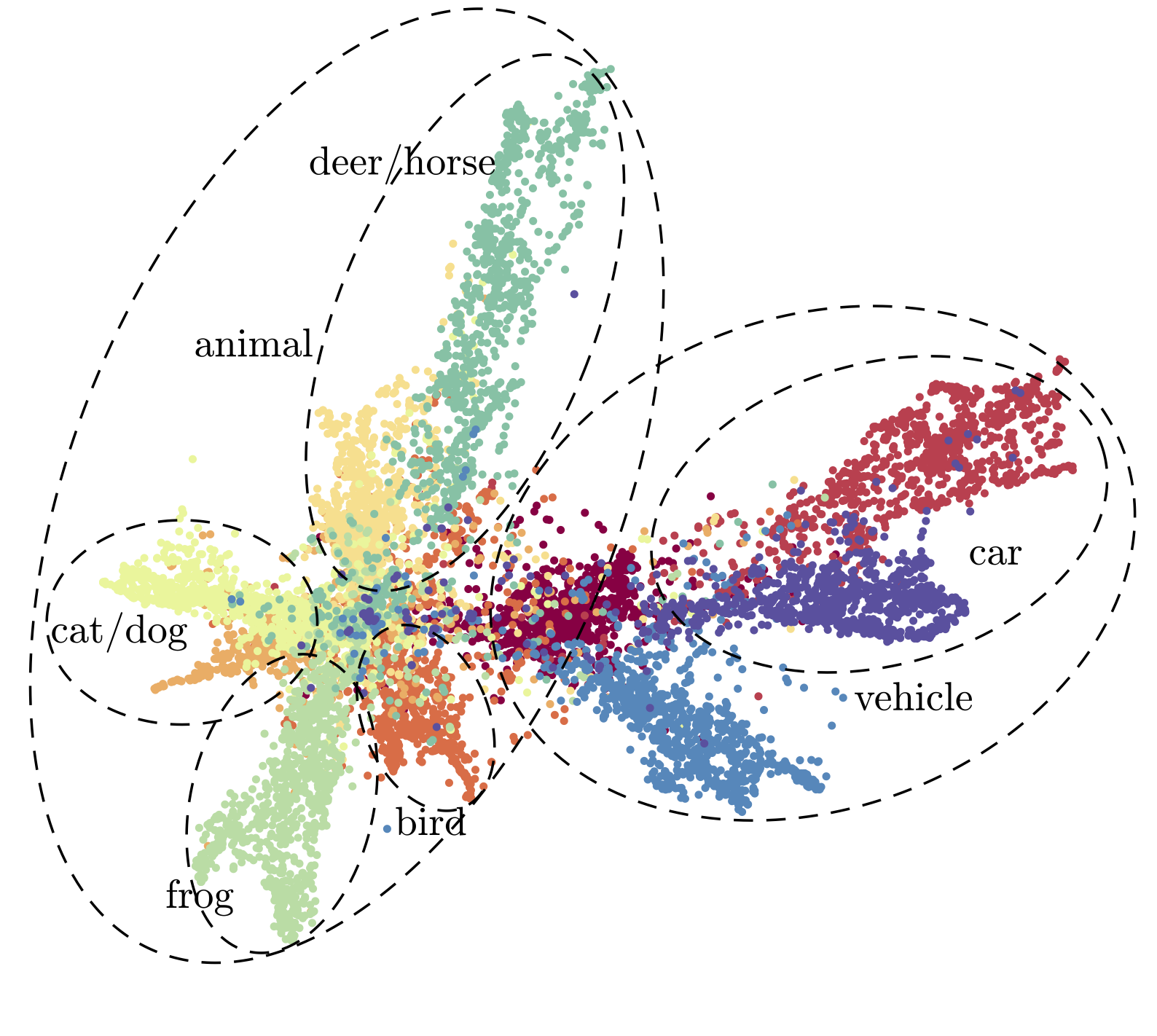}} \\
     (a) & (b)\\
    \end{tabular}
    }
    \caption{Visualization of the output of the first fully-connected layer of a convolutional network trained on the CIFAR-10 dataset (see the experiments section for details): (a) t-SNE, (b) TriMap. Each plot shows the embedding of the test set containing \num{10000} images. The values of nearest-neighbor accuracy and global score (introduced later) are shown as a pair \texttt{(AUC,GS)} on top of each plot. Note that t-SNE plot fails to present the underlying structure. In contrast, TriMap plot shows multiple hierarchies in the data: the two super-clusters corresponding to ``animal'' and ``vehicle'' as well as multiple smaller clusters are successfully uncovered. Note that a higher GS value for TriMap reflects this fact.}\label{fig:fc1}
    \end{center}
    \vspace{-0.5cm}
    \end{table}

We first introduce the \emph{global score},
a quantitative measure that reflects the closeness of a
given embedding to the PCA embedding (which is
optimal by means of preserving the data variance).
The purpose of this score is to measure the accuracy of an embedding in reflecting the overall placement of the clusters of points relative to their original representation in high-dimension. By design, PCA yields the highest global score among all the DR methods, and high values of global score indicate the efficacy of a DR method in reflecting the global structure.

Next, we introduce \emph{TriMap}, a DR method that focuses on preserving the global structure of the data in the embedding.
Pairwise (dis)similarities between points (used by the previous DR
methods) seem to be insufficient in capturing the global structure.
Instead, TriMap incorporates a higher order of structure to construct the embedding by means of \emph{triplets}:
\begin{quote}
\begin{center}
    $(i,j,k) \Leftrightarrow\,\, $\emph{point $i$ is closer to point $j$ than point $k$.}
    \end{center}
\end{quote}
The key idea behind TriMap stems from semi-supervised
metric learning~\citep{semi}:
Given an initial low-dimensional representation for the
data points, the triplet information from the high-dimensional representation of the points
is used to enhance the quality of the embedding.
Similarly, TriMap is initialized with the low dimensional
PCA embedding and this embedding is then modified using a set of carefully selected triplets
from the high-dimensional representation.

With an extensive set of experiments, we show that TriMap produces excellent results on a variety of real-world as well as synthetic datasets. We show that in many cases, TriMap outperforms all the competitor non-linear methods by means of the global score and provides comparable local accuracy. While being significantly faster than t-SNE, TriMap provides comparable runtime to UMAP and LargeVis while scaling drastically better to larger datasets. On the Character Font Images dataset of $\sim\!\! 1.7$M points, TriMap calculates the embedding in $\sim\!\! 1.3$ hours while LargeVis takes more than $3$ hours and UMAP exceeds the $12$ hours time limit. Our contributions can be summarized as follows:
\begin{itemize}
    \item We introduce a global score to quantify the quality of a low-dimensional embedding in reflecting the global structure of the high-dimensional data, such as placement of the clusters rather than the local neighborhood of individual points.
    \item We introduce TriMap, a fast dimensionality reduction method that provides embeddings of the data that are globally more accurate than other non-linear DR methods such as t-SNE, LargeVis, and UMAP.
    \item We provide an efficient implementation\footnote{\url{https://github.com/eamid}} of TriMap that can easily scale to millions of points on commodity hardware and outperforms the competing methods in terms of runtime. We also perform many large-scale experiments on various datasets to show the efficacy of TriMap in terms of DR performance measures and runtime. [Update] A JAX implementation of TriMap is available at:~\url{https://github.com/google/trimap}.
\end{itemize}

\section{A Measure of Global Accuracy}

Consider the {\sf S}-curve
dataset\footnote{\url{https://scikit-learn.org/stable/modules/generated/sklearn.datasets.make_s_curve.html}}
which consists of $\num{5000}$ points in 3-D uniformly
sampled from an {\sf S}-shaped manifold
(Figure~\ref{fig:scurve}.(a)). This dataset serves as a
paradigmatic problem for evaluating the performance of DR
methods. In Figure~\ref{fig:scurve}, we show the results of
2-D embeddings of the {\sf S}-curve dataset using t-SNE,
UMAP, TriMap, and PCA. The top of each graph
is labeled by the scoring pair (AUC, GS) where GS stands for global score (introduced below).
Note that both t-SNE and UMAP provide higher values of the
AUC score and locally preserve the continuity of the manifold.
However, they both fail to recover the global structure of
the {\sf S}-curve, which is naturally reflected in the PCA
embedding. On the other hand, our TriMap method (formally defined later) successfully recovers the structure of the {\sf S}-curve
by ``unveiling'' the curved shape of the manifold at both ends.
Overall, the 2-D TriMap embedding resembles the original 3-D representation
as much as possible.
Note also that GS is the only measure that can reflect the global accuracy of the embedding.

% \begin{figure*}
\begin{table}[t!]
\vspace{-1.2cm}
\setlength{\tabcolsep}{4pt}
\renewcommand{\tablename}{Figure}
\begin{center}
\resizebox{\textwidth}{!}{
\begin{tabular}{ccccc}
    {\footnotesize Original \texttt{(AUC,GS)}} & {\footnotesize t-SNE (\textbf{0.18}, 0.18)}& {\footnotesize UMAP (0.16, 0.13)} & {\footnotesize TriMap (0.15, \textbf{0.80})} & {\footnotesize PCA (0.03, 1.00)}\\[-1mm]
    \subfigure{\includegraphics[ width=0.24\textwidth]{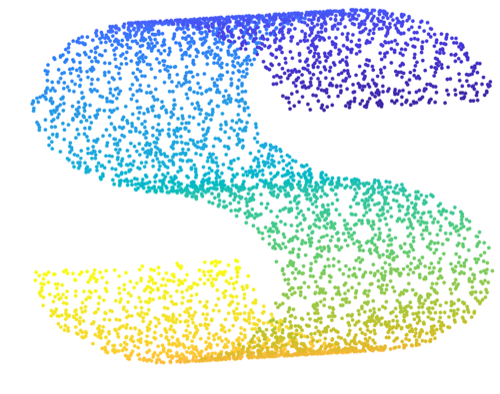}} &
    \subfigure{\includegraphics[ width=0.24\textwidth]{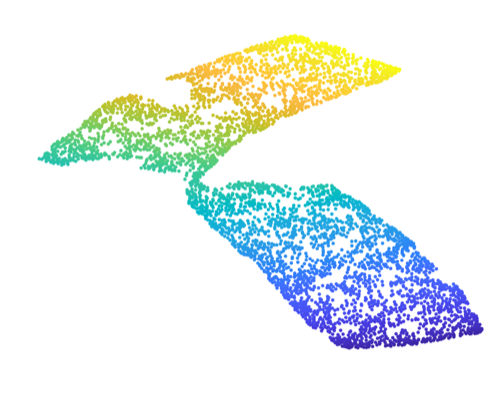}} &
    \subfigure{\includegraphics[ width=0.24\textwidth]{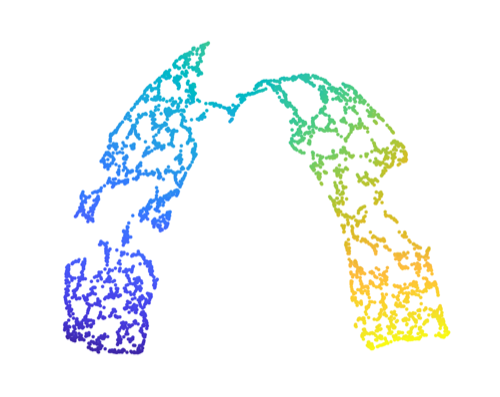}} &
    \subfigure{\includegraphics[ width=0.24\textwidth]{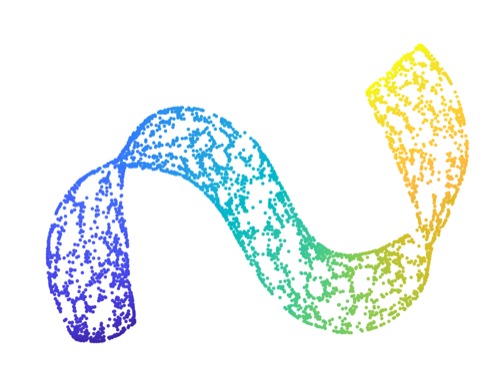}} &
    \subfigure{\includegraphics[ width=0.24\textwidth]{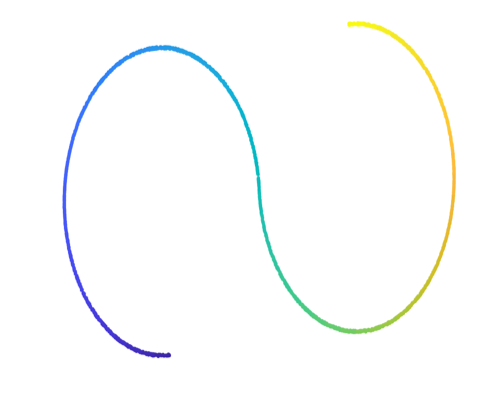}}\\
    (a) & (b) & (c) & (d) & (e)\\
    \end{tabular}
    }
    \caption{2-D Visualizations of the {\sf S}-curve dataset: (a) original dataset in 3-D, (b) t-SNE, (c) UMAP, (d) TriMap, and (e) PCA. The AUC and global score values for, respectively, measuring local and global accuracy, are shown in order as a pair (AUC, GS) for each embedding. Despite having higher AUC values, t-SNE and UMAP both fail to reflect the overall shape of the {\sf S}-curve. On the other hand, TriMap successfully unveils the underlying structure in the original dataset. GS is the only DR performance measure that can reflect this property.}
    \label{fig:scurve}
    \vspace{-0.35cm}
    \end{center}
    \end{table}
% \end{figure*}

The previous example indicates that the local measures of
DR performance (such as AUC) cannot reflect the global
accuracy of a low-dimensional embedding. In fact, the
low-granular structure of the data can only be estimated by
considering the global statistics of the dataset, as
regarded by the PCA method. PCA is a linear DR method that
projects the high-dimensional data onto the top-$\ldim$
orthogonal directions having the highest variance. In order
to calculate the mapping, PCA only considers the aggregate
statistics of the dataset rather than the local information
of each individual data point. As a result, PCA is
extremely well suited at retaining the \emph{global
structure} of the data, i.e., the overall shape of the dataset, placement of the clusters, and the existence of potential outliers. However, by focusing on the global structure, PCA loses much of the local information, such as the neighborhood structure of each data point.

Given a low-dimensional mapping produced by PCA, it is possible to calculate an optimal inverse mapping into the original high-dimensional space by means of minimizing the squared error. The optimal inverse map also corresponds to a linear mapping.\footnote{More specifically, mapping to low-dimension corresponds to projecting the data onto the top-$\ldim$ eigendirections of the data covariance matrix. The inverse mapping is induced by the transpose of the projection matrix.} In order to quantify the global accuracy of a DR result, we focus on the accuracy of the embedding in reflecting the global structure of the data similar to PCA. That is, we consider the minimum reconstruction error of the original dataset by means of a linear inverse map. Given $n$ data points $\{\x_i \in \RR^\hdim\}_{i=1}^n$, let $\X \in \RR^{\hdim \times n}$ denote the high-dimensional data matrix where the $i$-th column corresponds to $\x_i$. Similarly, let $\Y \in \RR^{\ldim \times n}$ denote the matrix of the low-dimensional embedding of the points $\{\y_i \in \RR^\ldim\}_{i=1}^n$. Without loss of generality, we assume both $\X$ and $\Y$ are centered. We define the \emph{Minimum Reconstruction Error (MRE)} from the embedding as
\[
\err(\Y|\, \X) \coloneqq \min_{\A \in \RR^{\hdim \times \ldim}} \Vert \X - \A \Y\Vert_{\text{F}}^2\, ,
\]
where $\Vert\,\cdot\,\Vert_{\text{F}}$ denotes the Frobenius norm.\footnote{The optimum value $\A^*$ for the MRE can be calculated efficiently as
\[
\A^* = \X\Y^\top(\Y\Y^\top)^{-1}\, .
\] This also handles possible rotation and scaling of the embedding.}
Note that PCA has the lowest possible MRE among all the DR methods. Thus, in order to obtain a normalized measure of global accuracy of a given embedding $\Y$ for a data $\X$, we define the \emph{global score (GS)} as
\[
\GS(\Y|\, \X) \coloneqq \exp\Big(-\frac{\err(\Y|\, \X) - \err_{\text{\tiny PCA}}}{\err_{\text{\tiny PCA}}}\Big) \in [0, 1]\, ,
\]
where $\err_{\text{\tiny PCA}} \coloneqq \err(\Y_{\text{\tiny PCA}}|\, \X)$ denotes the MRE achieved by the PCA embedding $\Y_{\text{\tiny PCA}}$ on the same dataset $\X$.
Note that $\GS(\Y_{\text{\tiny PCA}}|\, \X) = 1$ and we suggest that larger values of GS indicate a higher capacity of a DR method to reflect
the global structure of the data, as shown in the experiments.

\iffalse
\begin{table}[t!]
\vspace{-1.2cm}
\setlength{\tabcolsep}{4pt}
\renewcommand{\tablename}{Figure}
\begin{center}
\resizebox{\textwidth}{!}{
\begin{tabular}{cccc}
    {\footnotesize NIL \texttt{(NN\! =\! $0.906$,GS\,=\,$0.927$)}} & {\footnotesize $\gamma = 50$ \texttt{(NN\,=\,$0.935$,GS\,=\,$0.924$)}} & {\footnotesize $\gamma = 500$ \texttt{(NN\,=\,$0.940$,GS\,=\,$0.916$)}} & {\footnotesize $\gamma = 5000$ \texttt{(NN\,=\,$0.941$,GS\,=\,$0.908$)}}\\[-1mm]
    \subfigure{\includegraphics[ width=0.3\textwidth]{./figs/mnist_weights_false.png}} &
    \subfigure{\includegraphics[ width=0.3\textwidth]{./figs/mnist_weights_50.png}} &
    \subfigure{\fbox{\includegraphics[ width=0.3\textwidth]{./figs/mnist_weights_5000.png}}} &
    \subfigure{\includegraphics[ width=0.3\textwidth]{./figs/mnist_weights_500.png}}\\
    (a) & (b) & (c) & (d)\\
    \end{tabular}
    }
    \vspace{-0.2cm}
    \caption{\stkout{The Effect of the weight transformation on the MNIST dataset:
    (a) no weight transformation, (b) $\gamma = 50$, (c) $\gamma = 500$ (default),
    and (d) $\gamma = 5000$. The values of nearest neighbor accuracy and global score are shown as a tuple \texttt{(NN,GS)} on top of each figure. Larger values of $\gamma$ emphasizes more on the local accuracy rather than the global accuracy.} \Blue{[Update] This feature is deprecated and is replaced by a temperature parameter with a default value of \texttt{weight\_temp} = $0.5$.}}\label{fig:gamma-effect}
    \vspace{-0.4cm}
    \end{center}
    \end{table}
\fi

In the remainder of the paper, we use GS as the global performance measure.
Due to the high computational complexity for calculating
the trustworthiness-continuity and AUC scores for large data sets,
we use nearest-neighbors accuracy as the local measure of performance henceforth.

\section{The TriMap Method}

We now formally introduce the TriMap method. Recall that a triplet consists of three points $(i,j,k)$
where point $i$ is closer to point $j$ than point $k$.
TriMap chooses a subset $\T = \{(i,j,k)\}$ of triplets and assigns a weight $\w_{ijk} \geq 0$ for each triplet:
a higher value of $\w_{ijk}$ implies that the pair $(i,k)$
is located much farther than the pair $(i,j)$. We define the loss of the triplet $(i,j,k)$ as
\[
\loss_{ijk} \coloneqq \w_{ijk}\, \frac{s(\y_i, \y_k)}{s(\y_i, \y_j) + s(\y_i, \y_k)}\text{,\,\,\, where \,\,\,}
s(\y_i, \y_j) = \big(1 + \vert \y_i - \y_j \Vert^2\big)^{-1}\, ,
\]
\iffalse
\setcounter{figure}{3}
\begin{wrapfigure}{r}{0.3\textwidth}
\vspace{-0.95cm}
  \begin{center}
    \includegraphics[ width=0.24\textwidth]{./figs/gamma.png}
  \end{center}
  \vspace{-0.7cm}
  \caption{\stkout{$\gamma$-scaled log-transformation with
    different values of $\gamma$. The value NIL corresponds
    to no transformation.}} \label{fig:gamma}
%cant be fixed by adding vspace here
\end{wrapfigure}
\fi
is a similarity function between $\y_i$ and $\y_j$. The choice of $s$ is motivated by the good performance of Student t-distribution for similarities in  low-dimension in the t-SNE method. Note that the loss of the triplet $(i,j,k)$ approaches zero as $\Vert \y_i - \y_j\Vert$ decreases and $\Vert \y_i - \y_k\Vert$ increases.

We first develop the
weighing scheme for the triplets.
To reflect the relative similarities in high-dimension, we define the  weight of the triplet $(i,j,k)$ as
\[
\tilde{\w}_{ijk} = d_{ik}^2-d_{ij}^2\, \geq 0\, ,
\]
in which, $d_{ij}$ is any distance measure between $\x_i$ and $\x_j$ in high-dimension. As the default distance, we consider the squared Euclidean distance. We also apply the scaling introduced in~\citep{scaling},
\[
d_{ij}^2 = \frac{\Vert\x_i - \x_j \Vert^2}{\sigma_{ij}}\, ,
\]
where  $\sigma_{ij} = \sigma_i\, \sigma_j$ and $\sigma_i$ is set to the average Euclidean distance between $\x_i$ and the set of nearest-neighbors of $\x_i$ from $4$-th to $6$-th neighbors. This choice of $\sigma_{ij}$ adaptively adjusts the scaling based on the density of the data.

\iffalse
\sout{While the choice of weights $\tilde{\w}_{ijk}$ works well in practice, we adjust the weights further by applying a non-linear transformation that emphasizes the smaller weights. Expanding the values of small weights has the effect of placing the nearest-neighbors closer to the point and pushing the remaining points farther away, thus improving the local accuracy (as shown in Figure~\ref{fig:gamma-effect} and discussed later). The final value of the weight $\w_{ijk}$ is obtained by applying the $\gamma$-scaled log-transformation (see Figure~\ref{fig:gamma}),}
\[
\stkout{\w_{ijk} = \zeta_\gamma\big(\frac{\tilde{\w}_{ijk}}{\mathcal{W}} + \delta\big) \text{ \,\,\, where \,\,\,}\zeta_\gamma(u) \coloneqq \log\big(1 + \gamma\, u\big)\,,}
\]
 \sout{in which $\mathcal{W} = \max_{(i^{\scriptscriptstyle \prime}\!,j^{\scriptscriptstyle \prime}\!,k^{\scriptscriptstyle \prime})\in \T} \tilde{\w}_{i^{\scriptscriptstyle \prime}\!j^{\scriptscriptstyle \prime}\!k^{\scriptscriptstyle \prime}}$, \, $\gamma > 0$ is a scaling factor, and $\delta$ is a small constant. We use $\gamma = 500$ and $\delta=10^{-4}$ in all our experiments.}
 \fi
 [Update] We shift the final weights by subtracting the minimum weight value calculated over all triplets and applying a tempered log transformation,
 \[
 \w_{ijk} = \log_t\big(1 + \tilde{\w}_{ijk} - \w_{\min}\big)\, ,
 \]
 where $\w_{\min} = \min_{(i^{\scriptscriptstyle \prime}\!,j^{\scriptscriptstyle \prime}\!,k^{\scriptscriptstyle \prime})\in \T} \tilde{\w}_{i^{\scriptscriptstyle \prime}\!j^{\scriptscriptstyle \prime}\!k^{\scriptscriptstyle \prime}}$, and
 \[
 \log_t (u) = \frac{1}{1-t}\, \big(u^{1-t} - 1\big)\, , \,\, t \neq 1\, ,
 \]
 is called the tempered logarithm~\citep{texp1}. The $\log_t$ transformation smoothens the weights and prevents the triplets with large weights from dominating the total loss. Note that the limit case $t\rightarrow 1$ recovers the standard $\log$. We use $t=0.5$ as the default value.

\setcounter{table}{2}
\begin{table}[t!]
\vspace{-1.3cm}
\setlength{\tabcolsep}{4pt}
\renewcommand{\tablename}{Figure}
\begin{center}
\resizebox{\textwidth}{!}{
\begin{tabular}{ccccc}
    {\footnotesize $\vert \T\vert\!=\!3n$ \texttt{(NN\,=\,$0.360$,GS\,=\,$0.885$)}} & {\footnotesize $\vert \T\vert\!=\!10n$ \texttt{(NN\,=\,$0.787$,GS\,=\,$0.908$)}} & {\footnotesize $\vert \T\vert\!=\!55n$ \texttt{(NN\,=\,$0.940$,GS\,=\,$0.916$)}} & {\footnotesize $\vert \T\vert\!=\!210n$ \texttt{(NN\,=\,$0.941$,GS\,=\,$0.923$)}} & {\footnotesize $\vert \T\vert\!=\!820n$ \texttt{(NN\,=\,$0.929$,GS\,=\,$0.927$)}}\\[-1mm]
    \subfigure{\includegraphics[ width=0.34\textwidth]{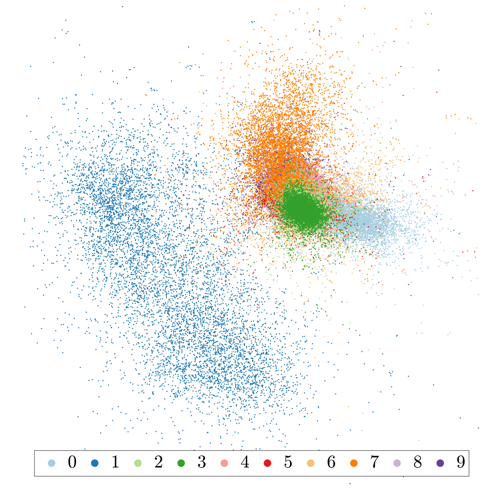}} &
    \subfigure{\includegraphics[ width=0.34\textwidth]{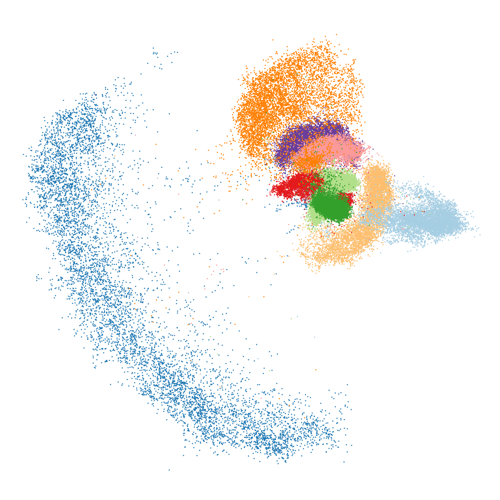}} &
    \subfigure{\fbox{\includegraphics[ width=0.34\textwidth]{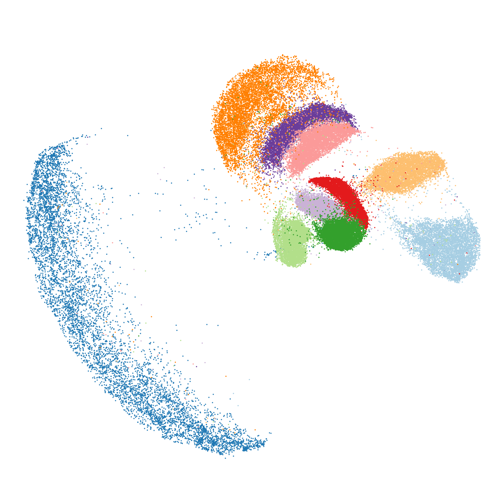}}} &
    \subfigure{\includegraphics[ width=0.34\textwidth]{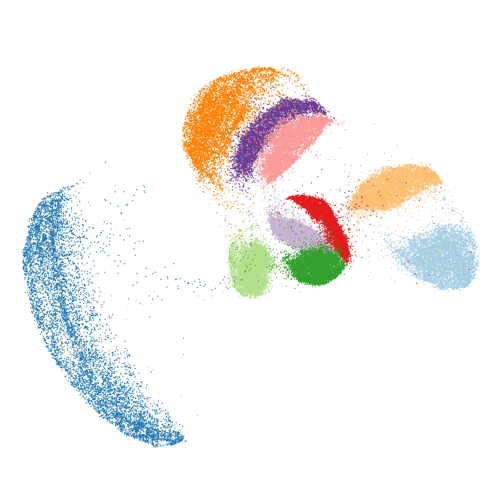}} &
    \subfigure{\includegraphics[ width=0.34\textwidth]{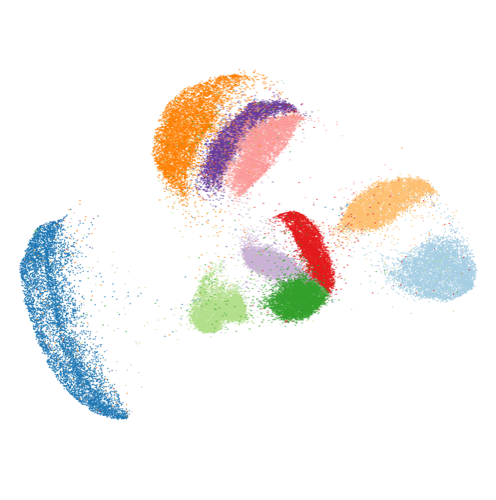}}
    \\
    (a) & (b) & (c) & (d) & (e)\\
    \end{tabular}
    \vspace{-0.2cm}
    }
    \caption{Effect of changing the number of triplets on the quality of the embeddings of the MNIST dataset. We consider  $(m, m^\prime, r) = c\times(2,1,1)$ for: (a) $c=1$, (b) $c=2$, (c) $c=5$ (default) [Update] The new default values are $(m, m^\prime, r) = (12, 4, 3)$, (d) $c=10$, and (e) $c=20$. The values of nearest neighbor accuracy and global score are shown as a tuple \texttt{(NN,GS)} on top of each figure. The quality of embedding does not improve significantly after adding a certain number of triplets.}\label{fig:num_triplets}
    \vspace{-0.35cm}
    \end{center}
    \end{table}

To construct the embedding, we consider a small subset of all possible triplets $(i,j,k)$ for which the closer point $j$ belongs to the set of nearest-neighbors of the point $i$ and the farther point $k$
is among the points that are more distant from $i$ than $j$,
chosen uniformly at random.
For each point we consider its $m = 10$ nearest neighbors and sample $m^\prime = 5$ triplets per nearest-neighbor. This yields $m\times m^\prime = 50$ nearest-neighbor triplets per point. In addition, we also add $r = 5$ random triplets $(i,j,k)$ per each point $i$ where $j$ and $k$ are sampled uniformly at random and their order is possibly switched based on their nearness to $i$. This yields $m\times m^\prime + r = 55$ triplets per point in total. ([Update] The new default parameters are set to $m = 12,\, m^\prime=4,$ and $r=3$.) Thus, the overall complexity of the optimization step is linear in number of points $n$. The computational complexity is dominated by the nearest-neighbor search, which is shared among all the recent methods such as t-SNE, LargeVis, and UMAP. We use ANNOY for the approximate nearest-neighbor search\footnote{\url{https://github.com/spotify/annoy}. [Update] The JAX implementation uses PyNNDescent: \url{https://github.com/lmcinnes/pynndescent}. } which is based on random projection trees.

\newcolumntype{C}[1]{>{\centering\arraybackslash}p{#1}}\setcounter{table}{0}
\newcolumntype{R}[1]{>{\raggedleft\arraybackslash}p{#1}}
\begin{table}[t]
\setlength{\tabcolsep}{9pt}
% \vspace{-1.3cm}
\begin{center}
\resizebox{0.8\textwidth}{!}{
\begin{tabular}{m{2.3cm} C{0.5cm} C{0.5cm} c C{0.5cm} C{0.5cm} c C{0.5cm}  C{0.5cm} c C{0.5cm}  C{0.5cm}}
\toprule
\multicolumn{1}{l}{Dataset} &
\multicolumn{2}{c}{t-SNE} & &
\multicolumn{2}{c}{UMAP} & &
\multicolumn{2}{c}{TriMap} & &
\multicolumn{2}{c}{PCA}\\
\cline{2-12}
& \texttt{PCC} & \texttt{GS} & & \texttt{PCC} & \texttt{GS} & & \texttt{PCC} & \texttt{GS} & & \texttt{PCC} & \texttt{GS}\\
\midrule
 COIL-20 &  0.56 & 0.63 && 0.30 & 0.52 && 0.50 & 0.69 && 0.88 & 1.00\\\hline
  USPS &  0.70 & 0.89  && 0.88 & 0.90 && 0.85 & 0.93 && 0.91 & 1.00\\\hline
Epileptic S. &  0.69 & 0.91  && 0.82 & 0.90 && 0.91 & 0.91 && 0.84 & 1.00\\\hline
Tabula M. &  0.56 & 0.05  && 0.30 & 0.09 && 0.82 & 0.17 && 0.93 & 1.00\\\hline
MNIST &  0.62 & 0.90  && 0.79 & 0.91 && 0.77 & 0.92 && 0.83 & 1.00\\\hline
Fashion MNIST &  0.66 & 0.53  && 0.90 & 0.75 && 0.92 & 0.87 && 0.97 & 1.00\\
% {\small \textbf{Tabula Muris} (54K)} &  0.88 & 0.88  & 0.56 & 0.56 & 0.69 & 0.69 & 0.30 & 0.30 & 0.49 & 0.49\\\hline
% {\small \textbf{MNIST} (70K)} &  0.88 & 0.88  & 0.56 & 0.56 & 0.69 & 0.69 & 0.30 & 0.30 & 0.49 & 0.49\\\hline
%  {\small \textbf{Fashion MNIST} (70K)} &  0.88 & 0.88  & 0.56 & 0.56 & 0.69 & 0.69 & 0.30 & 0.30 & 0.49 & 0.49\\\hline
%  {\small \textbf{360+K Lyrics} ($\sim$360K)} &  0.88 & 0.88  & 0.56 & 0.56 & 0.69 & 0.69 & 0.30 & 0.30 & 0.49 & 0.49\\\hline
% {\small \textbf{Covertype} ($\sim$581K)} &  0.88 & 0.88  & 0.56 & 0.56 & 0.69 & 0.69 & 0.30 & 0.30 & 0.49 & 0.49\\\hline
%  {\small\textbf{RCV1} (800K)} &  0.88 & 0.88  & 0.56 & 0.56 & 0.69 & 0.69 & 0.30 & 0.30 & 0.49 & 0.49\\
 \bottomrule
\end{tabular}
}
\vspace{-0.1cm}
\caption{Comparison of the Pearson correlation coefficient (PCC) and global score (GS) values for different method. Overall, TriMap yields higher scores than t-SNE. Notice the strong correlation between PCC and GS.}\label{tab:gs}
     \end{center}
% 	\vspace{-5mm}
\end{table}

While a random initialization for the embedding also works well in practice, we initialize the embedding to the PCA solution $\Y_{\text{\tiny PCA}}$ (scaled by a small constant value for better convergence). The PCA initialization for TriMap allows faster convergence while preserving much of the global structure discovered by PCA. Note that the other DR methods such as t-SNE are extremely sensitive to the initialization and do not converge well with any initial solution other than small random initialization around the origin.

We define the final loss as the sum of the losses of the sampled triplets in $\T$
\[
\vspace{-0.07cm}
\loss_{\text{\tiny TriMap}} = \sum_{(i,j,k) \in \T} \loss_{ijk}\, .
\]
The loss is minimized using the full-batch gradient descent with momentum using the delta-bar-delta method. In all our experiments, we perform $400$ iterations with the value of momentum parameter equal to $0.5$ during the first $250$ iterations and $0.8$ afterwards.

\newcolumntype{C}[1]{>{\centering\arraybackslash}p{#1}}\setcounter{table}{0}
\newcolumntype{R}[1]{>{\raggedleft\arraybackslash}p{#1}}
\setcounter{table}{1}
\begin{table}[t!]
\setlength{\tabcolsep}{9pt}
\vspace{-0.3cm}
\begin{center}
\resizebox{0.85\textwidth}{!}{
\begin{tabular}{m{4.5cm} C{1.2cm} C{1.2cm}  C{1.2cm} C{1.2cm} R{1.5cm}}
\toprule
{\small Dataset (size)} & {\small t-SNE} & {\small LargeVis} & {\small UMAP} & {\small TriMap} & {\small Speedup} \\ \midrule
 {\small \textbf{COIL-20} (\num{1440})} &  00:00:08 & 00:05:51  & 00:00:04   &  \textbf{00:00:02} & \multicolumn{1}{r}{2.00$\times$}\\\hline
 {\small \textbf{USPS} (11K)} &  00:02:02   &  00:06:12 &    00:00:12   &  \textbf{00:00:11} & \multicolumn{1}{r}{1.10$\times$}\\\hline
 {\small\textbf{Epileptic Seizure} (11.5K)} &  00:03:11 & 00:06:17 & 00:00:15   &  \textbf{00:00:12} & \multicolumn{1}{r}{1.25$\times$}\\\hline
 {\small \textbf{20 Newsgroup} (18K)} &  00:05:34   & 00:06:57 & 00:00:26   &  \textbf{00:00:21} & \multicolumn{1}{r}{1.24$\times$}\\\hline
 {\small \textbf{Tabula Muris} (54K)} &  00:17:32 & 00:09:29  & 00:01:12   &  \textbf{00:01:06} & \multicolumn{1}{r}{2.00$\times$}\\\hline
 {\small \textbf{MNIST} (70K)} &  00:20:38  & 00:11:29 & \textbf{00:01:15}   &  00:01:23 & \multicolumn{1}{r}{0.90$\times$}\\\hline
 {\small \textbf{Fashion MNIST} (70K)} &  00:19:10 & 00:11:04 & \textbf{00:01:18}   &  00:01:24 & \multicolumn{1}{r}{0.93$\times$}\\\hline
 {\small \textbf{TV News} ($\sim$129K)} &  00:38:59   &  00:16:26 &    00:02:57   &  \textbf{00:02:45} & \multicolumn{1}{r}{1.07$\times$}\\\hline
 {\small \textbf{360+K Lyrics} ($\sim$360K)} &  08:50:49   & 00:44:16 & 00:25:23   &  \textbf{00:13:49} & \multicolumn{1}{r}{1.84$\times$}\\\hline
 {\small \textbf{Covertype} ($\sim$581K)} &  --   & 00:44:54 & 02:59:41 &  \textbf{00:24:42} & \multicolumn{1}{r}{1.82$\times$}\\\hline
 {\small\textbf{RCV1} (800K)} &  --   & 01:34:38 & 04:55:53 & \textbf{00:36:59} & \multicolumn{1}{r}{2.56$\times$}\\\hline
 {\small \textbf{Character Font Images} ($\sim$1.7M)} &  -- & 03:16:19    & -- &  \textbf{01:17:50} & \multicolumn{1}{r}{2.52$\times$}\\\hline
 {\small \textbf{KDDCup99} ($\sim$4.9M)} &  --  & -- & -- &  \textbf{04:17:01} & \multicolumn{1}{c}{\,\,\,\,\,\,\,--}\\\hline
 %{\small \textbf{SUSY} (5M)} &  -- & --  & -- & -- \\\hline
 {\small \textbf{HIGGS} (11M)} &  -- & --  & -- &  \textbf{10:08:36} & \multicolumn{1}{c}{\,\,\,\,\,\,\,--}
 \\\bottomrule
\end{tabular}
}
\vspace{-0.1cm}
\caption{Runtime of the methods in \texttt{hh:mm:ss} format on single machine with 2.6 GHz Intel Core i5 CPU and 16 GB of memory. We limit the runtime of each method to 12 hours. Also, UMAP  runs out of memory on datasets larger than $\sim$4M points.}\label{tab:runtime}
     \end{center}
	\vspace{-5mm}
\end{table}

Finally, note that there exist connections between TriMap
and a number of \emph{triplet (aka ordinal) embedding}
methods such as t-STE~\citep{ste}. The triplet embedding
methods have been developed for a different setting where
the goal is to find an embedding based on a given
pre-specified set of triplets obtained from human
evaluators (or some form of implicit feedback). For
instance, t-STE maximizes the sum of log of the
satisfaction probabilities of the triplets to calculate the
embedding. It is worth mentioning that TriMap is a DR
method that is designed to sample the informative triplets
from the high-dimensional representation of a set of points
and assign weights to these triplets to reflect the
relative similarities of these points. Although TriMap can
also be used for the triplet embedding task, we only focus
on the DR results.~\footnote{Also, the embeddings obtained
by simply applying these methods to our set of  sampled
triplets are quite subpar (see the MNIST result
in~\citep{ste}) and are not shown here.}

\setcounter{figure}{3}
\begin{figure}[t!]
\vspace{-1.4cm}
    \centering
    \subfigure{\includegraphics[ width=0.24\textwidth]{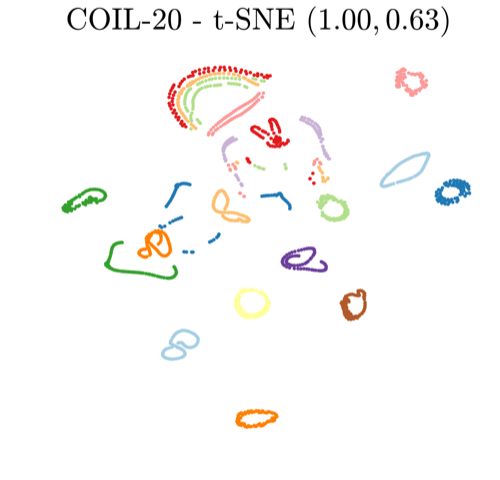}}
    \subfigure{\includegraphics[ width=0.24\textwidth]{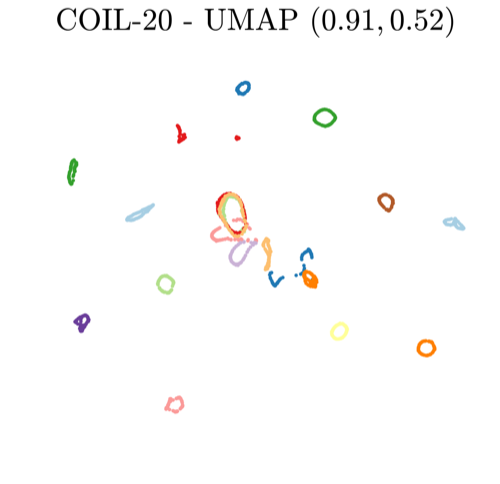}}
    \subfigure{\includegraphics[ width=0.24\textwidth]{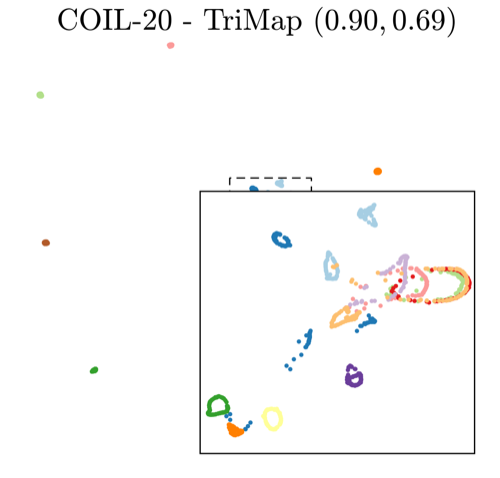}}
    \subfigure{\includegraphics[ width=0.24\textwidth]{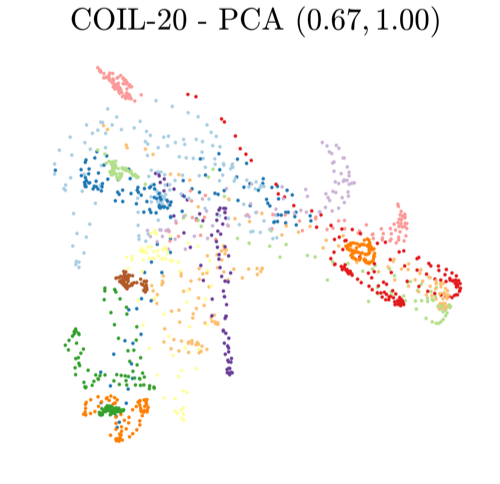}}\hfill\\[-3mm]
    \subfigure{\includegraphics[ width=0.24\textwidth]{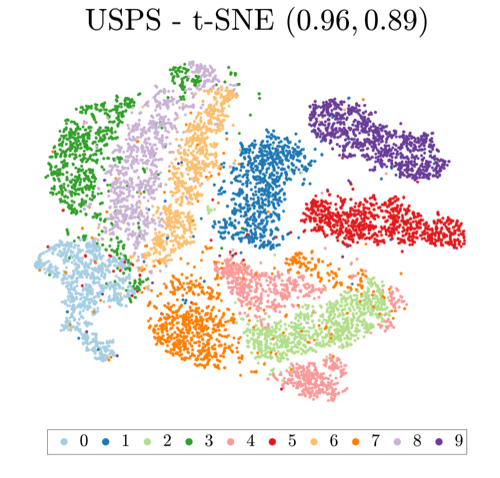}}
    \subfigure{\includegraphics[ width=0.24\textwidth]{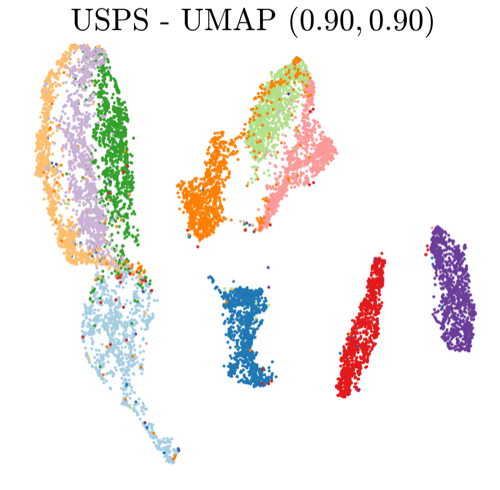}}
    \subfigure{\includegraphics[ width=0.24\textwidth]{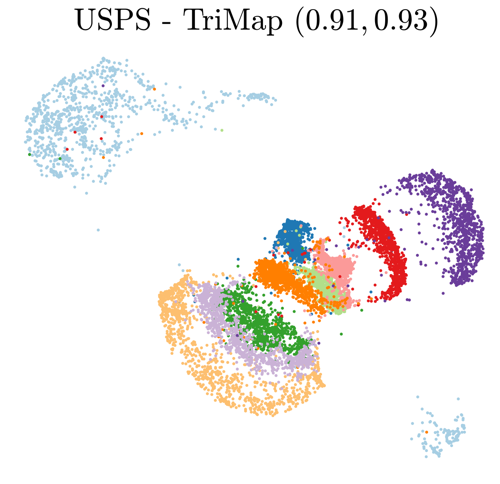}}
    \subfigure{\includegraphics[ width=0.24\textwidth]{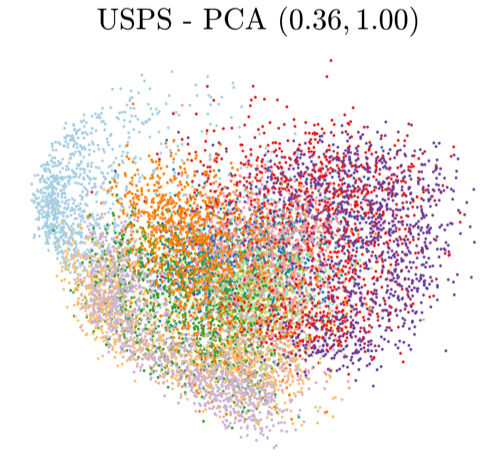}}\hfill\\[-3mm]
    \subfigure{\includegraphics[ width=0.24\textwidth]{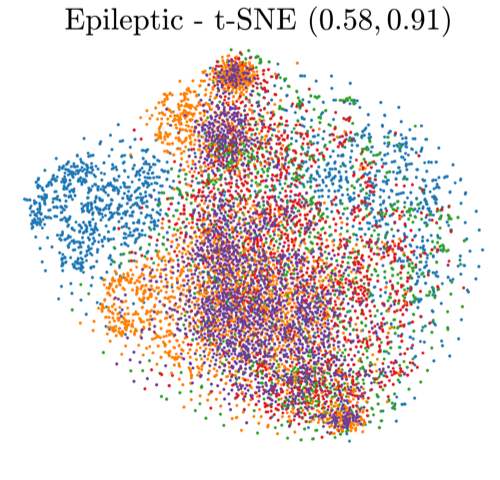}}
    \subfigure{\includegraphics[ width=0.24\textwidth]{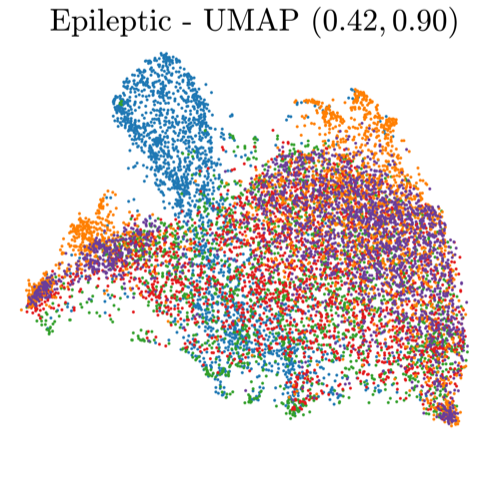}}
    \subfigure{\includegraphics[ width=0.24\textwidth]{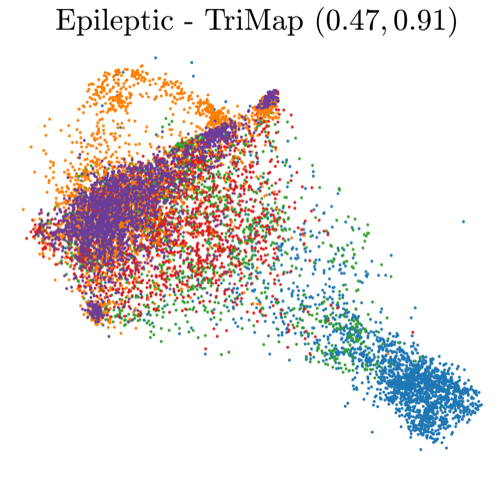}}
    \subfigure{\includegraphics[ width=0.24\textwidth]{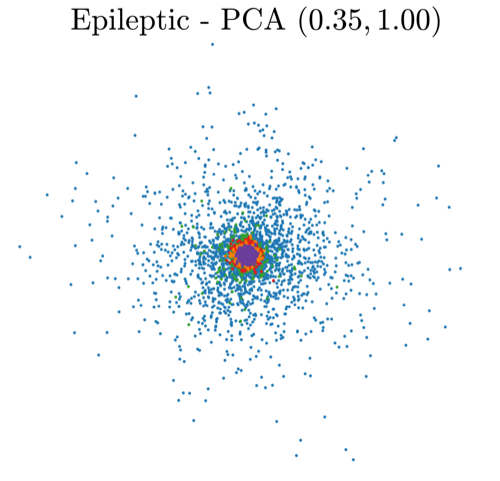}}\hfill\\[-3mm]
    \subfigure{\includegraphics[ width=0.24\textwidth]{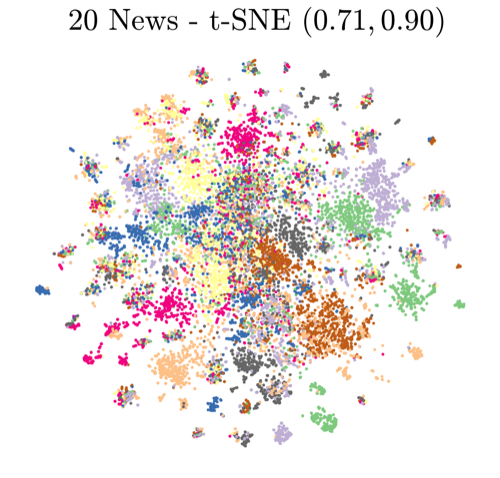}}
    \subfigure{\includegraphics[ width=0.24\textwidth]{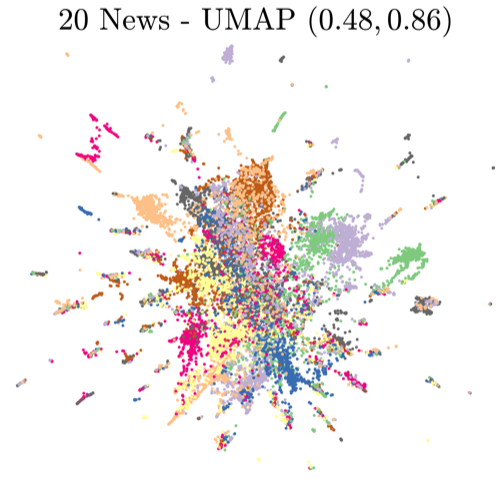}}
    \subfigure{\includegraphics[ width=0.24\textwidth]{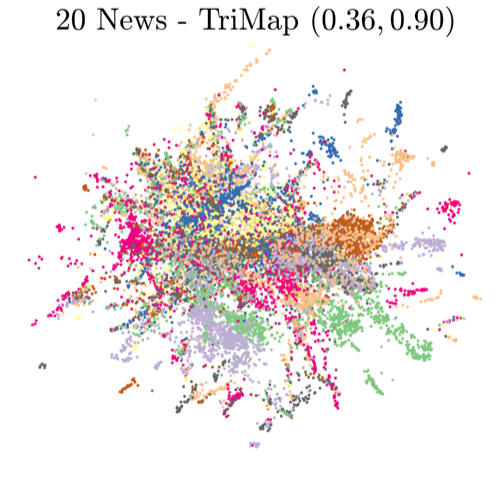}}
    \subfigure{\includegraphics[ width=0.24\textwidth]{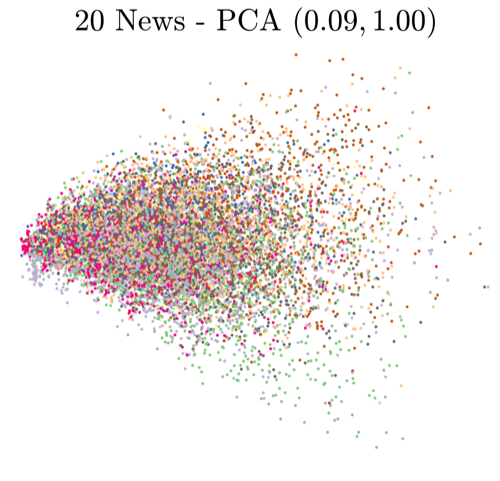}}\hfill\\[-3mm]
    \subfigure{\includegraphics[ width=0.24\textwidth]{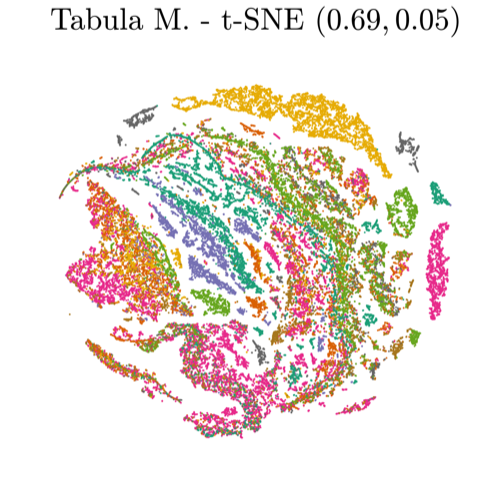}}
    \subfigure{\includegraphics[ width=0.24\textwidth]{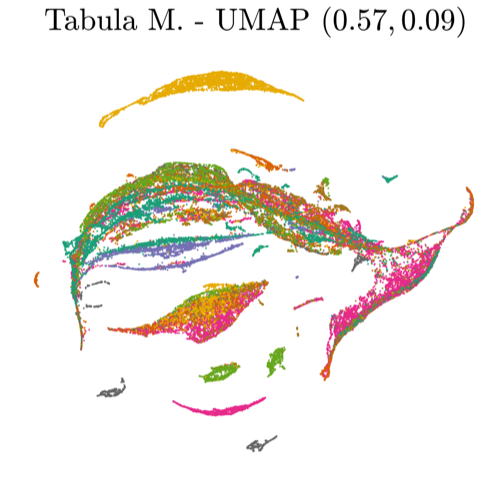}}
    \subfigure{\includegraphics[ width=0.24\textwidth]{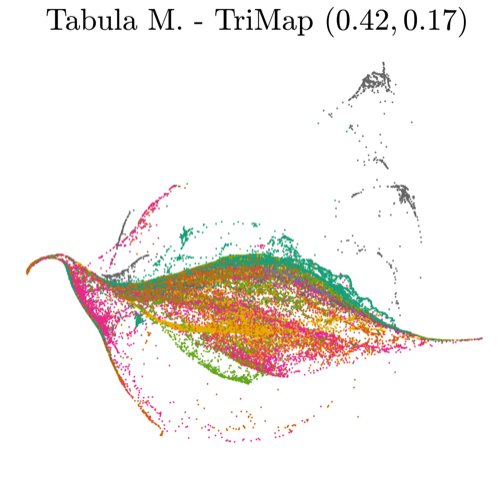}}
    \subfigure{\includegraphics[ width=0.24\textwidth]{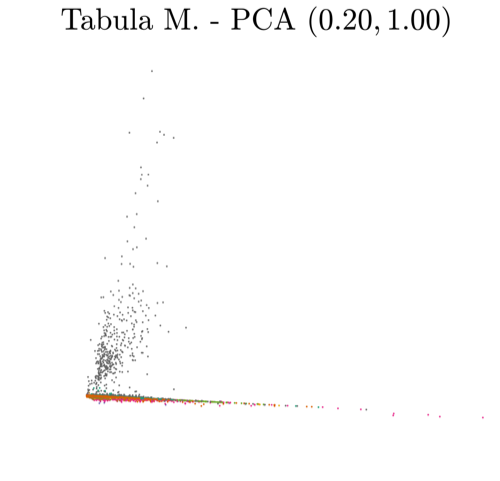}}\hfill\\
    \vspace{-0.3cm}
    \caption{Visualizations of different datasets using t-SNE, UMAP, TriMap, and PCA. Each row corresponds to one dataset and each column represents one method. The values of nearest neighbor accuracy and global score are shown as a pair \texttt{(NN,GS)} on top of each figure.}
    \label{fig:results}
    \vspace{-0.5cm}
\end{figure}

\subsection{Effect of Different Parameters}
We briefly discuss the effect of different parameters, namely the total number of triplets $\vert \T\vert$ and the $\gamma$-scaled log-transformation, on the quality of the embedding. TriMap is particularly robust to the number of sampled triplets for constructing the embedding. This can be explained by the high amount of redundancy in the triplets (the triplets $(i,j,k)$ and $(i,j,k^\prime)$ convey the same information if $k$ and $k^\prime$ are nearest neighbors and also mapped nearby). In Figure~\ref{fig:num_triplets} we consider various values for $m$, $m^\prime$, and $r$ for the MNIST dataset while fixing the remaining parameters.
In fact, using a large number of triplets can sometimes introduce overhead and require a larger number of iterations to converge.

\iffalse
\stkout{A more important parameter is $\gamma$ which controls the trade-off between the local and global accuracy. Larger values of $\gamma$ increases the relative importance of triplets with smaller weights. This causes the method to focus on the nearest-neighbor points rather than the points that are far away, thus improving the local accuracy. On the other hand, improving the local accuracy can impair global accuracy. In Figure~\ref{fig:gamma-effect}, we plot the $\gamma$-scaled log-transformation for various $\gamma$ values and illustrate the results on MNIST without the log-transformation as well as the results with different $\gamma$ values. For larger values of $\gamma$, the clusters tend to become more compressed, and as a result, the nearest-neighbor accuracy is improved. On the other hand, the global score starts to decrease for larger $\gamma$ values.}% We use $\gamma = 500$ as the default value.
\vspace{-0.2cm}
\fi
\section{Experiments}
\vspace{-0.1cm}

In this section, we apply TriMap on a set of real-world as well as synthetic datasets and compare the results to t-SNE, LargeVis, UMAP, and PCA methods. The datasets used in our experiments are listed in Table~\ref{tab:runtime} and a short description is given in the appendix. All experiments are conducted on a single machine with 2.6 GHz Intel Core i5 CPU and 16 GB of memory. We limit the runtime of each algorithm to 12 hours. For implementations, we use the default \texttt{sklearn} implementation for t-SNE and the official implementations of LargeVis and UMAP provided by the authors\footnote{\url{https://github.com/lferry007/LargeVis}}\footnote{\url{https://github.com/lmcinnes/umap}}. Due to lack of space, we provide the comparison to the LargeVis results as well as additional TriMap results on the larger datasets in the appendix.

In order to have a fair comparison, we use the default parameter values for all methods, including ours ($m=10$, $m^\prime = 5$, $r=5$, \stkout{$\gamma = 500$}, and 400 iterations). ([Update] The new default parameters are set to $m=12$, $m^\prime = 4$, $r=3$, $t=0.5$, and $\gamma$-scaled log-transformation has been removed.) Also, to reduce the overhead induced by the dimensionality of the data in the nearest-neighbor search step, we reduce the number of dimensions of the dataset to 100 if necessary, using the PCA method. To evaluate the local performance, we show the nearest-neighbor accuracy of each result. We also show the GS as a measure of global performance. The performance measures are shown on top of each figure as a pair \texttt{(NN, GS)}.

\subsection{Global Accuracy}
We first validate the efficacy of the GS measure by drawing a comparison to the recently proposed Pearson Correlation Coefficient (PCC)~\citep{freedman2007statistics}. To calculate PCC, we consider random subsets of size \num{10000} from different datasets and perform $k$-means clustering in the high-dimensional space with $k=100$. Next, we perform the clustering in the low-dimensional embedding on the same subset of points and calculate the score between the distance matrices of the cluster centers in the high-dimension and low-dimension using the Mantel test. We repeat the process $100$ times using different sub-samples of the data. The results are shown in Table~\ref{tab:gs}. As can be seen, PCC admits a high correlation with the GS. Overall, PCC and GS are the highest for PCA (GS is equal to $1$ by definition). On the other hand, TriMap yields excellent global performance and results in much higher PCC and GS scores compared to t-SNE.

\subsection{Visualization of Standard Datasets}
The visualizations of the datasets using TriMap as well as the other competing methods are shown in Figure~\ref{fig:results} and~\ref{fig:more-results}. We provide a zoomed-in snippet over the main figure for some results to provide a more detailed illustration. Overall, TriMap preserves the underlying global structure of the data better than the other competing methods. This is reflected by the larger GS values for TriMap as well as visually comparing the embeddings to the PCA result.  For example, TriMap recovers the continuous structure of the TV news dataset and separates the remaining outliers in the data, which are also identified by the PCA method. This can be verified by comparing the placement of an example outlier point, marked with a red $\bm{\times}$, by the different methods: TriMap shows this point among other outliers, whereas t-SNE and UMAP fail to uncover this information.  Also, the global score of TriMap on this dataset is much higher than the other methods. Further discussion is given in the appendix.

\setcounter{figure}{3}
\begin{figure*}
\vspace{-1.4cm}
    \centering
    \subfigure{\includegraphics[ width=0.24\textwidth]{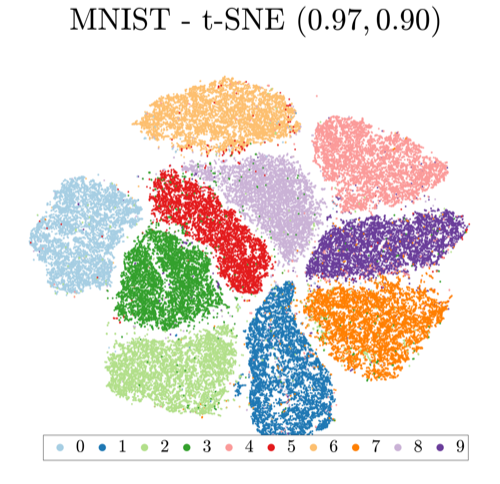}}
    \subfigure{\includegraphics[ width=0.24\textwidth]{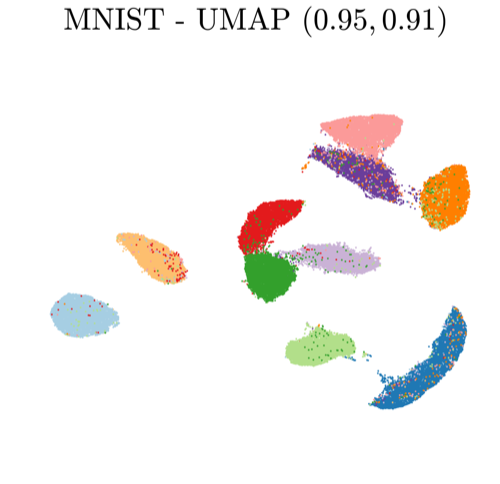}}
    \subfigure{\includegraphics[ width=0.24\textwidth]{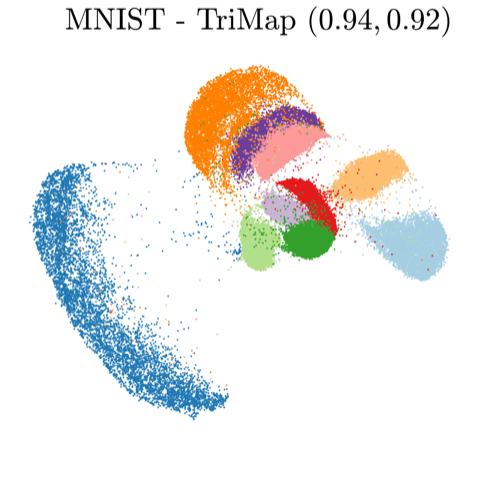}}
    \subfigure{\includegraphics[ width=0.24\textwidth]{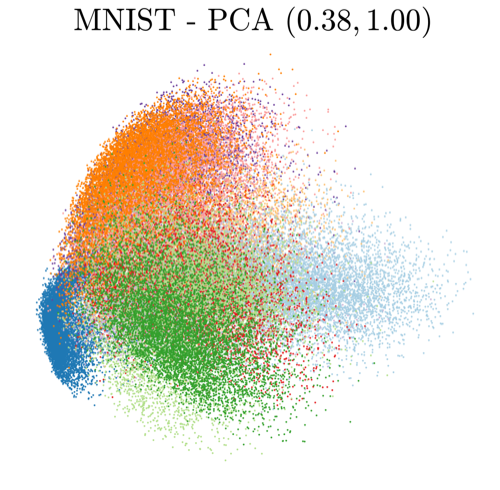}}\hfill\\[-3mm]
    \subfigure{\includegraphics[ width=0.24\textwidth]{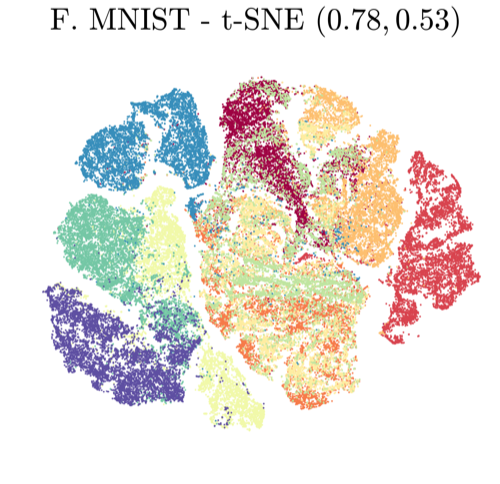}}
    \subfigure{\includegraphics[ width=0.24\textwidth]{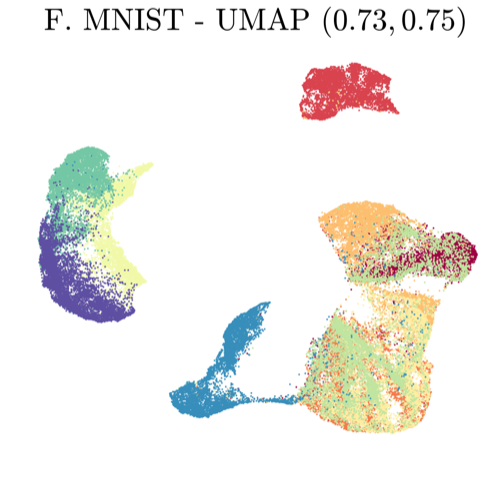}}
    \subfigure{\includegraphics[ width=0.24\textwidth]{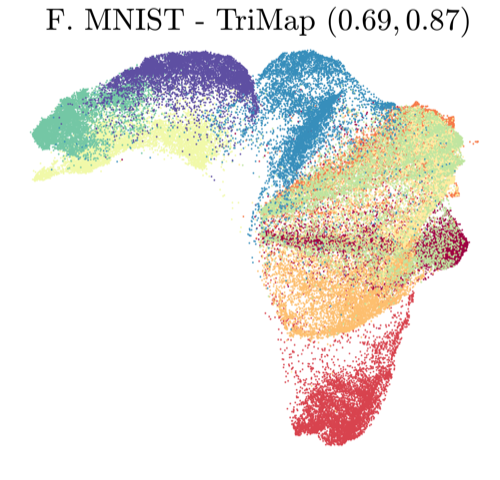}}
    \subfigure{\includegraphics[ width=0.24\textwidth]{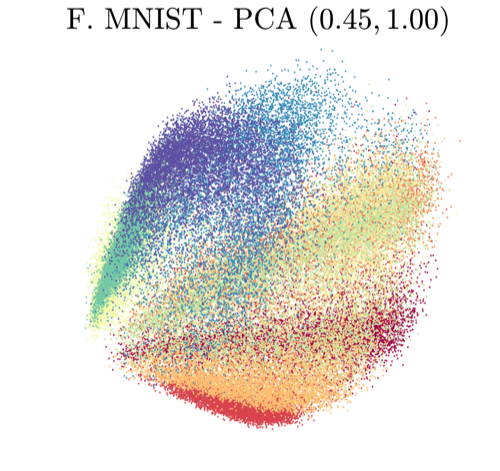}}\hfill\\[-3mm]
    \subfigure{\includegraphics[ width=0.24\textwidth]{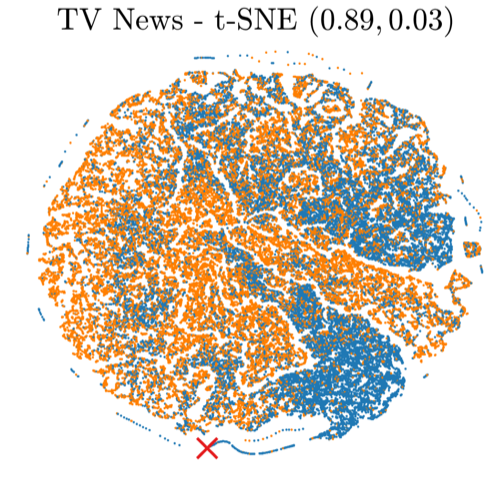}}
    \subfigure{\includegraphics[ width=0.24\textwidth]{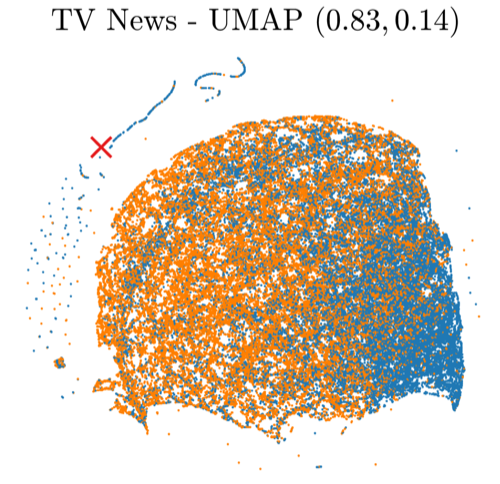}}
    \subfigure{\includegraphics[ width=0.24\textwidth]{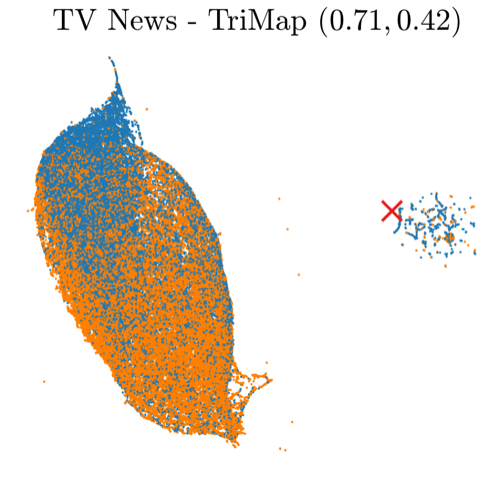}}
    \subfigure{\includegraphics[ width=0.24\textwidth]{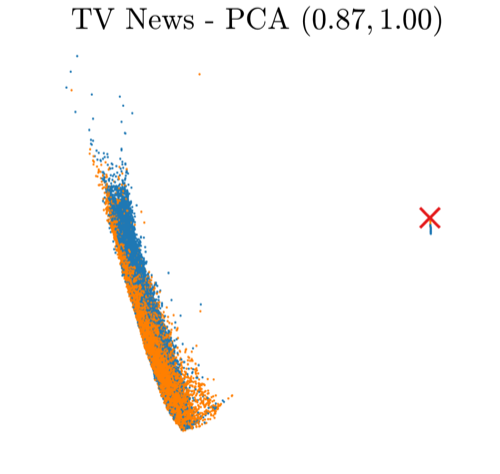}}\hfill\\[-3mm]
    \subfigure{\includegraphics[ width=0.24\textwidth]{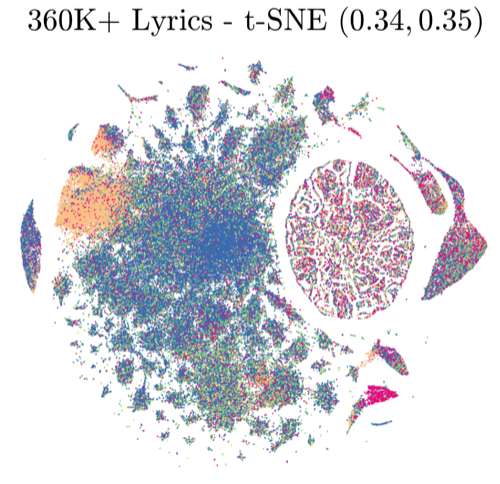}}
    \subfigure{\includegraphics[ width=0.24\textwidth]{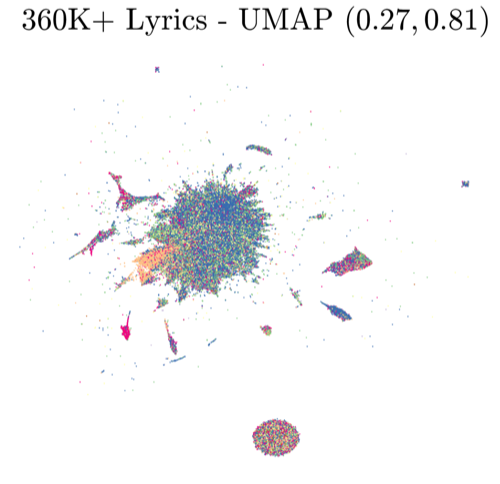}}
    \subfigure{\includegraphics[ width=0.24\textwidth]{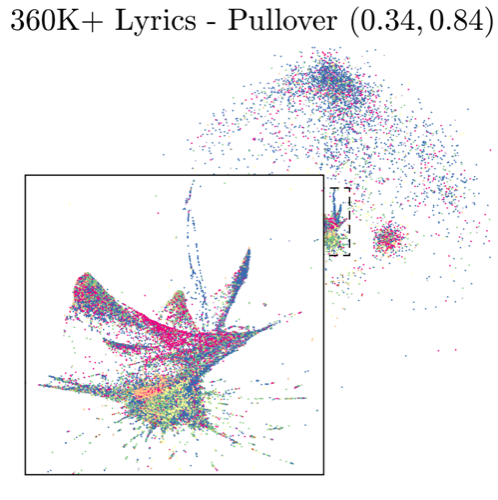}}
    \subfigure{\includegraphics[ width=0.24\textwidth]{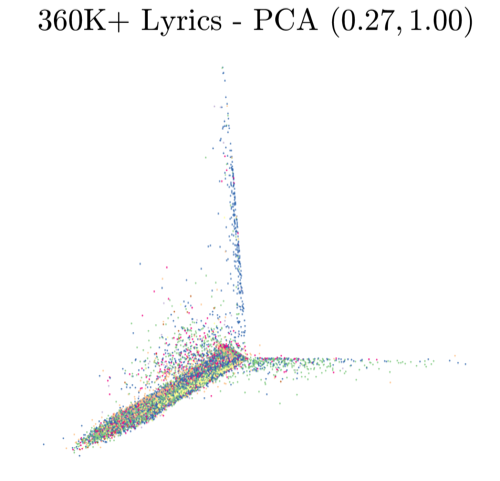}}\hfill\\
    \vspace{-0.3cm}
    \caption{Visualizations of different datasets (continued) using t-SNE, UMAP, TriMap, and PCA. Each row corresponds to one dataset and each column represents one method. The values of nearest neighbor accuracy and global score are shown as a tuple \texttt{(NN,GS)} on top of each figure.}
    \vspace{-0.2cm}
\end{figure*}

\setcounter{figure}{4}
\begin{figure}[t!]
\vspace{-1.4cm}
    \centering
    \subfigure{\includegraphics[ width=0.32\textwidth]{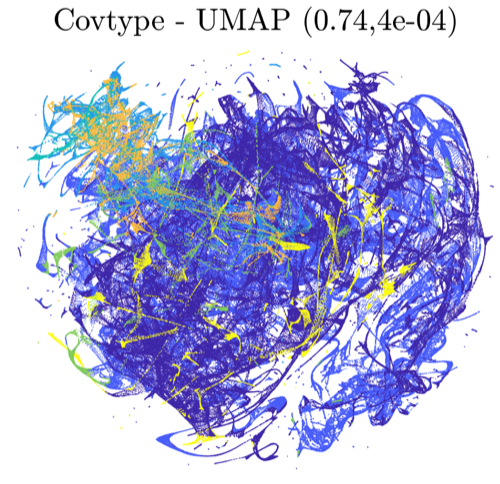}}
    \subfigure{\includegraphics[ width=0.32\textwidth]{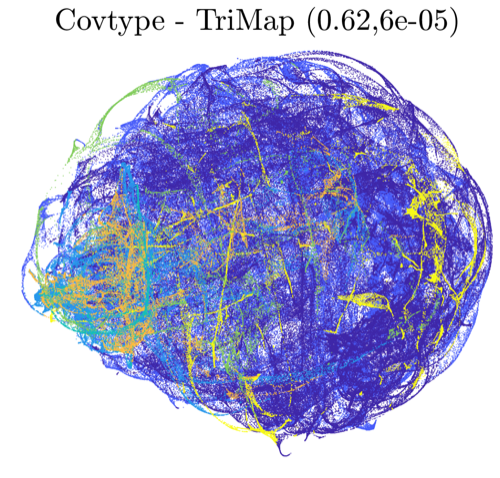}}
    \subfigure{\includegraphics[ width=0.32\textwidth]{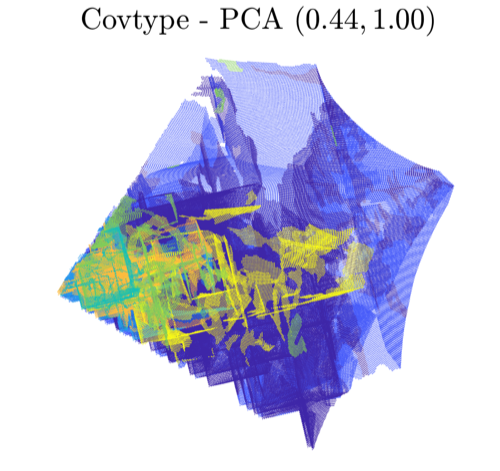}}\hfill\\[-3mm]
    \subfigure{\includegraphics[ width=0.32\textwidth]{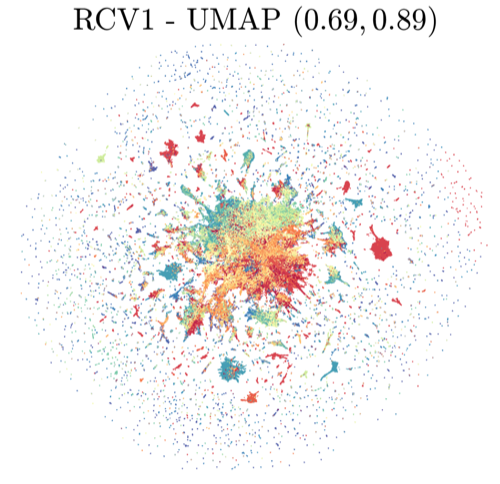}}
    \subfigure{\includegraphics[ width=0.32\textwidth]{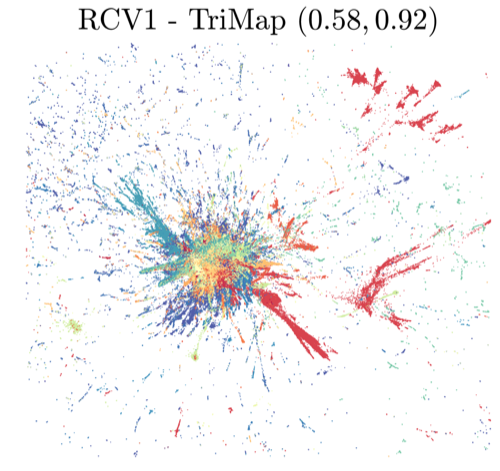}}
    \subfigure{\includegraphics[ width=0.32\textwidth]{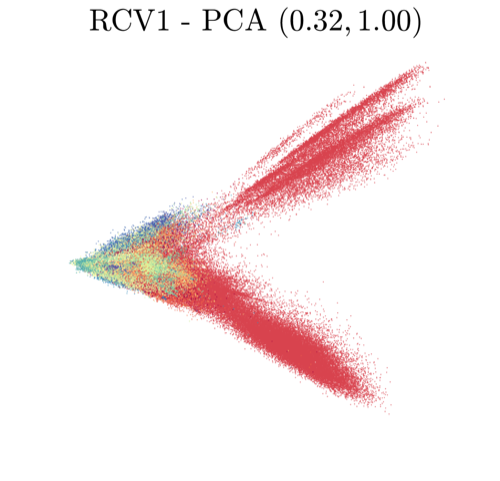}}\hfill\\[-3mm]
    \subfigure{\includegraphics[ width=0.32\textwidth]{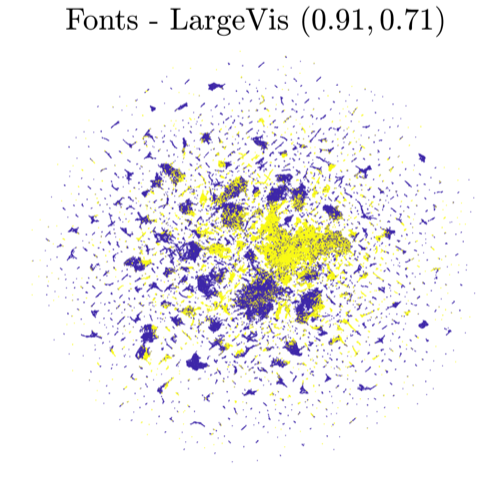}}
    \subfigure{\includegraphics[ width=0.32\textwidth]{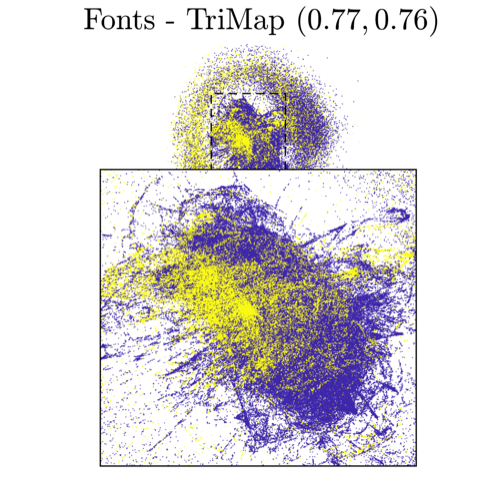}}
    \subfigure{\includegraphics[ width=0.326\textwidth]{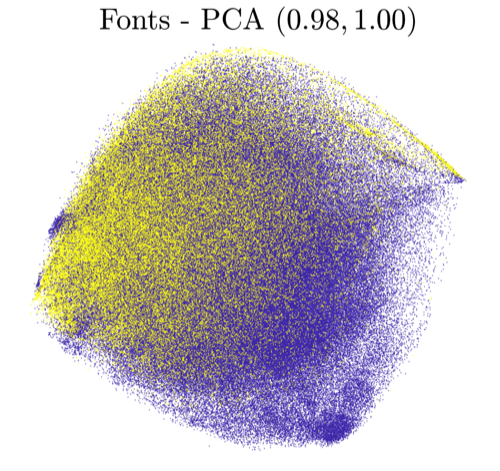}}\hfill\\
    % \vspace{-0.2cm}
    \caption{Visualizations of Covertype and RCV1 datasets using UMAP, TriMap, and PCA, and visualizations of the Character Font Images dataset using LargeVis, TriMap, and PCA. The values of nearest neighbor accuracy and global score are shown as a tuple \texttt{(NN,GS)} on top of each figure.}\label{fig:more-results}
    \vspace{-0.3cm}
\end{figure}

% \vspace{-0.2cm}
\subsection{Runtime}
The runtime of the methods are provided in Table~\ref{tab:runtime} in the \texttt{hh:mm:ss} format. We limit the runtime of each method to 12 hours.  As can be seen from the results, TriMap provides excellent runtime and outperforms all the other methods in most cases. Also, TriMap easily scales to millions of points while the other methods exceed the time limit or run out of memory. For instance, UMAP  causes an out of memory error for datasets larger than $\sim$4M points.

\subsection{Visualization of Neural Networks}

We visualize the different embeddings of the test set (\num{10000} points) from the CIFAR-10 dataset when passed through different layers of a convolutional neural network. Specifically, the network has $3$ convolutional layers, each sized $64$, followed by $3$ fully-connected layers of size $1024$, $512$, and $10$, respectively. We use ReLU activations and apply batch normalization and dropout with a keep probability of $0.75$ during training. We perform $25$ epochs of batch size $512$ using Adagrad optimizer on the \num{60000} point training set. The trained network achieves $84\%$ test accuracy. Note that the goal here is not to achieve the benchmark test accuracy but rather to observe the separation of clusters corresponding to different classes when passed through the network and to detect the highly misclassified examples. Figure~\ref{fig:fc1} shows the visualization of the output of the first fully-connected layer of the network on the test set using t-SNE and TriMap. The values of nearest-neighbor accuracy and global score are shown as a pair \texttt{(AUC,GS)} on top of each plot. Note that t-SNE shows the different clusters correctly but fails to present the overall underlying structure. In contrast, TriMap plot shows multiple hierarchies in the data: the two super-clusters corresponding to ``animal'' and ``vehicle'' as well as multiple smaller clusters (e.g. ``deer/horse'', ``cat/dog'', etc.) are successfully uncovered. Note that a higher GS value for TriMap reflects this fact.

We visualize the output of the second fully-connected layer in Figure~\ref{fig:fc2} (top). We can see that the clusters get separated better after passing through another layer of non-linearity. We also color the embedding based on the probability of the correct class, predicted by the network in Figure~\ref{fig:fc2} (bottom). In the plot, the color ``blue'' indicates a high probability for the correct (i.e., label) class while ``red'' corresponds to a misclassification. In both t-SNE and TriMap embeddings, the highly misclassified points are concentrated in the center of the embedding. However, the TriMap embedding reveals some examples that are also misclassified with high confidence but are far away from the center. We show an example of such points (marked with $\bm{\times}$) from the class ``dog'' which has been misclassified with high confidence as class ``'horse'' (with probability $\sim 1$). The same point is placed much closer to the center in the t-SNE plot, therefore not shown as a clear outlier. The actual image of the point overlayed in the TriMap plot shows the high resemblance of the example to a ``horse''.

\setcounter{table}{5}
\begin{table}[t!]
\vspace{-1.2cm}
\setlength{\tabcolsep}{1pt}
\renewcommand{\tablename}{Figure}
\begin{center}
\resizebox{0.9\textwidth}{!}{
\begin{tabular}{cc}
     {\footnotesize FC2 - t-SNE \texttt{(NN\,=\,$\bm{0.87}$,GS\,=\,$0.61$)}} & {\footnotesize FC2 - TriMap \texttt{(NN\,=\,$0.77$,GS\,=\,$\bm{0.84}$)}}\\[-1mm]
    \subfigure{\includegraphics[ width=0.45\textwidth]{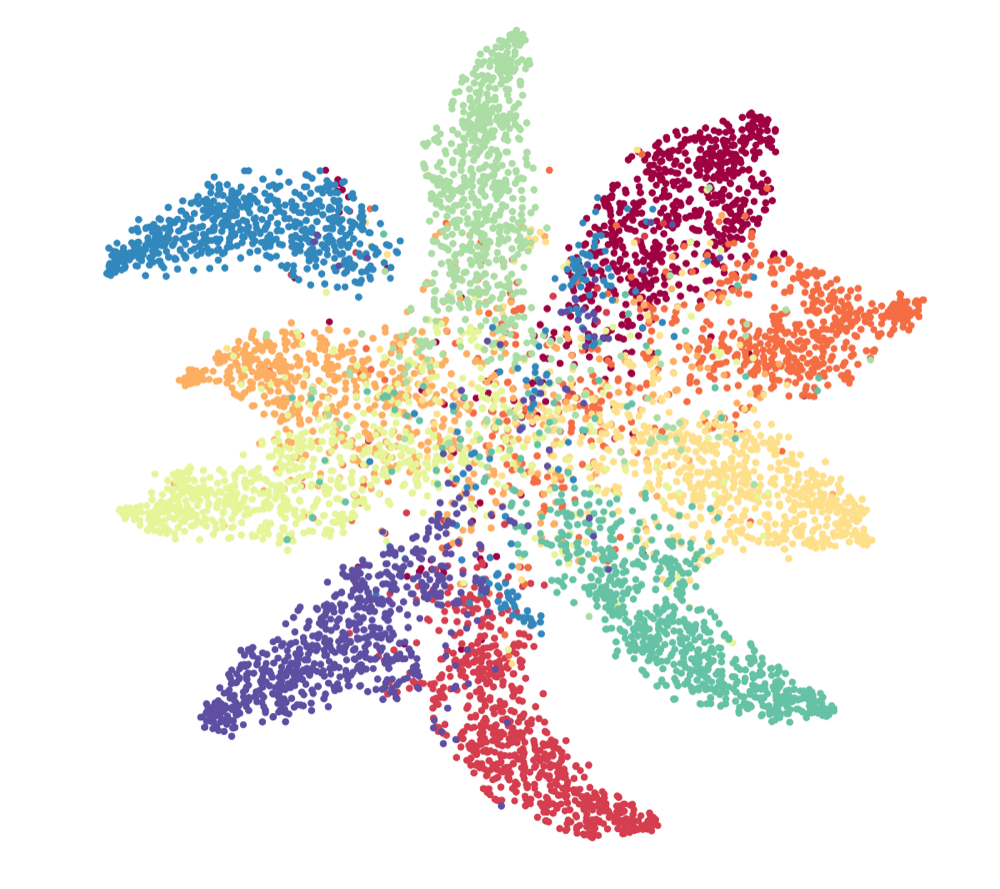}} &
    \subfigure{\includegraphics[ width=0.45\textwidth]{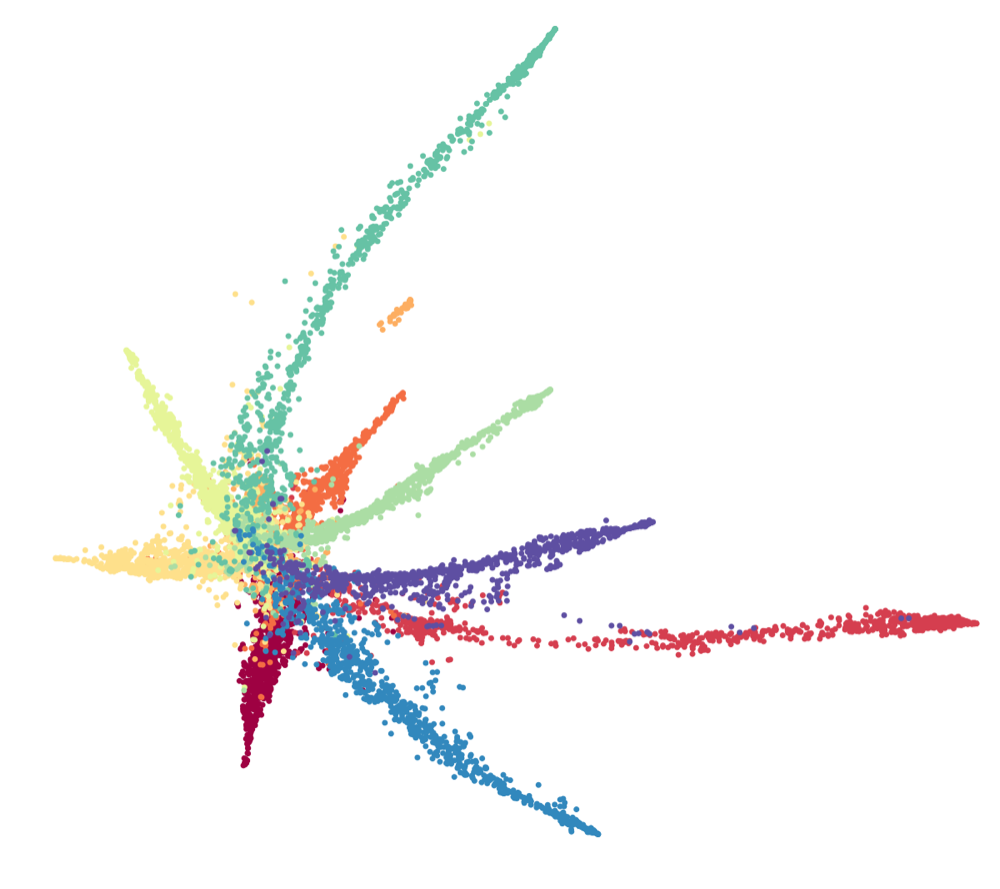}} \\
    \subfigure{\includegraphics[ width=0.45\textwidth]{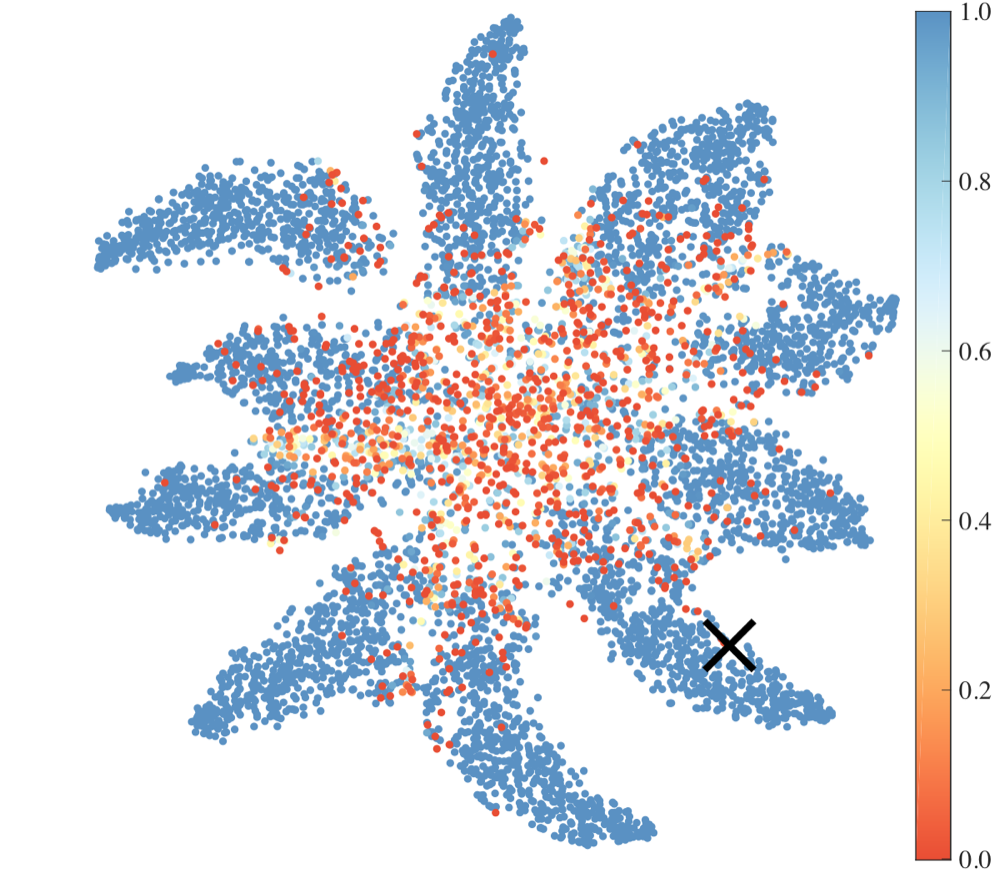}} &
    \subfigure{\includegraphics[ width=0.45\textwidth]{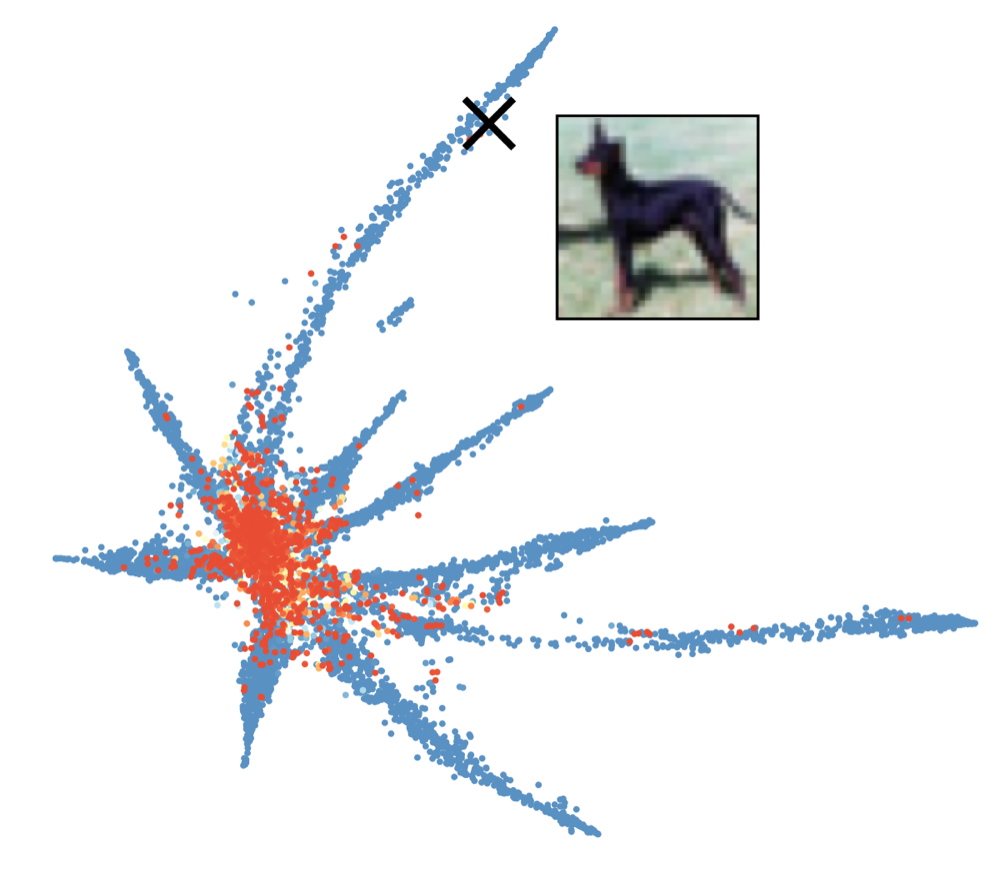}} \\
     (a) & (b)\\
    \end{tabular}
    \vspace{-0.2cm}
    }
    \caption{Visualization of the output of the second fully-connected layer: (a) t-SNE, (b) TriMap. We can observe that the clusters get separated better after passing through another layer of non-linearity. The bottom plot is colored based on the probability of the correct class. An example of a highly misclassified example from the class ``dog'' (classified as ``'horse'') is marked with $\bm{\times}$. The point is placed much closer to the center in the t-SNE plot, therefore not shown as a clear outlier. The actual image of the point is overlayed in the TriMap plot. The higher value of GS reflects the better global accuracy of TriMap.}\label{fig:fc2}
    \end{center}
    \vspace{-0.6cm}
    \end{table}

\section{Conclusion and Future Work}
TriMap is a fast and efficient method that can be easily
applied to large datasets. While TriMap is extremely
effective for uncovering the global structure of the data,
other methods such as t-SNE can provide additional insight
about the local neighborhood of individual points. As
a future research direction, we consider using pairwise
constraints along with triplet constraints to improve the
local accuracy. The current implementation of TriMap
utilizes a single core. Parallel implementation of the
method that can exploit multiple cores is another future direction. Furthermore, the global
accuracy is measured in terms of the global score, which is
based on the assumption that linear projection obtained by PCA is globally optimal. While
our global score can provide insight into the global
accuracy of the embedding in many cases, it appears to be
ineffective when the data is highly non-linear or contains
a large number of outliers. Developing non-linear and more
robust global performance measures could significantly
improve the assessment of the DR results and provide guidelines for developing more accurate
DR techniques.

% \section{TODO}
% \begin{itemize}
% \item UMAP is slow when number of connected componnets is large
% \item Shortcomings of the method
% \item future work how to combine pairwise with triplets
% \item better nearest neighbor search
% \item GS is not the best measure
% \item Faster implementation using multiple cores
% \item GS is not be best especially when there is noise in the data, better global scores
% \end{itemize}

\bibliography{refs}
\bibliographystyle{iclr2020_conference}

\newpage
\appendix
\section{Datasets}

The datasets used in the experiments are listed below. All datasets are publicly available online and a download link is provided.

\begin{itemize}
\setlength{\parskip}{3.5pt}
\setlength{\itemsep}{0pt}
    \item \textbf{COIL-20}\footnote{\url{http://www.cs.columbia.edu/CAVE/software/softlib/coil-20.php}} (\num{1440}): gray-scale images of $20$ objects in uniformly sampled orientations ($5$ degrees of rotation, $72$ images per object). Each image is pre-processed by having the background removed and cropped into size $128\times 128$.
    \item \textbf{USPS}\footnote{\url{https://www.kaggle.com/bistaumanga/usps-dataset}} (11K): images of handwritten digits (0--9) of size $16\times 16$.
    \item \textbf{Epileptic Seizure}\footnote{\label{fn:uci}\url{http://archive.ics.uci.edu/ml/index.php}} (11.5K): EEG signal recordings of brain activity for seizure recognition. It contains $178$-dimensional vectors belonging to $5$ categories.
    \item \textbf{20 Newsgroup}$^\text{\ref*{fn:uci}}$ (18K): newsgroup posts categorized into 20 topics. We use a TF-IDF representation of the words in each document as the features.
    \item \textbf{Tabula Muris}\footnote{\url{https://tabula-muris.ds.czbiohub.org/}} ($\sim$54K): single cell transcriptome data from the mouse from 20 organs.
    \item \textbf{MNIST}\footnote{\url{http://yann.lecun.com/exdb/mnist/}} (70K): images of handwritten digits (0--9) of size $28 \times 28$.
    \item \textbf{Fashion MNIST}\footnote{\url{https://github.com/zalandoresearch/fashion-mnist}} (70K): gray-scale images of clothing items such as t-shirt, pullover, bag, etc. of size $28 \times 28$.
    \item \textbf{TV News}$^\text{\ref*{fn:uci}}$ ($\sim$129K): audio-visual features from TV news broadcast categorized into commercial and non-commercial.
    \item \textbf{360K+ Lyrics}\footnote{\url{https://www.kaggle.com/gyani95/380000-lyrics-from-metrolyrics}} ($\sim$362K): lyrics of songs from $12$ different genres. We group similar genres together (metal-rock, R\&B-pop, etc.) to form 7 groups. We use the TF-IDF representation of the words in the song as the features.
    \item \textbf{Covertype}$^\text{\ref*{fn:uci}}$ ($\sim$581K): cartographic features for forest cover type prediction.
    \item \textbf{RCV1}\footnote{\url{https://scikit-learn.org/0.18/datasets/rcv1.html}} (800K): Reuters Corpus Volume I archive of  categorized newswire stories.
    \item \textbf{Character Font Images}$^\text{\ref*{fn:uci}}$ ($\sim$1.7M): images of character from scanned and computer generated fonts.
    \item \textbf{KDDCup99}$^\text{\ref*{fn:uci}}$ ($\sim$4.9M): computer network intrusion detection.
    %\item \textbf{SUSY} (5M): super-symmetric particle recognition from a background process.
    \item \textbf{HIGGS}$^\text{\ref*{fn:uci}}$ (11M): Higgs bosons recognition from a background process.
\end{itemize}

\section{More Visualizations}
We compare the results of TriMap to LargeVis in Figure~\ref{fig:results-lv} and~\ref{fig:more-results-lv}. We also provide more visualizations obtained using TriMap in Figure~\ref{fig:higgs}.

\setcounter{figure}{6}
\begin{figure}[t!]
\vspace{-1.6cm}
    \centering
    \subfigure{\includegraphics[ width=0.31\textwidth]{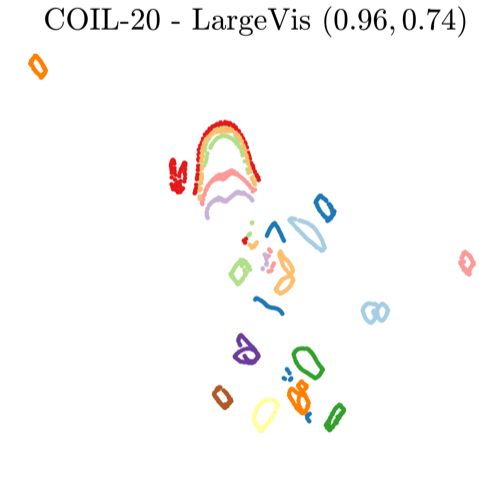}}
    \subfigure{\includegraphics[ width=0.31\textwidth]{./figs/coil20_trimap.png}}
    \subfigure{\includegraphics[ width=0.31\textwidth]{./figs/coil20_pca.png}}\hfill\\[-1mm]
    \subfigure{\includegraphics[ width=0.31\textwidth]{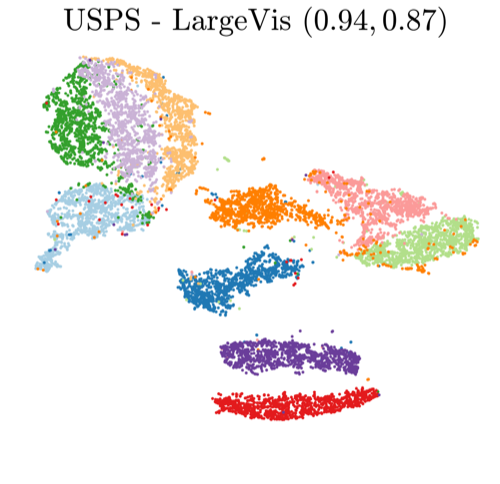}}
    \subfigure{\includegraphics[ width=0.31\textwidth]{./figs/usps_trimap.png}}
    \subfigure{\includegraphics[ width=0.31\textwidth]{./figs/usps_pca.png}}\hfill\\[-1mm]
    \subfigure{\includegraphics[ width=0.31\textwidth]{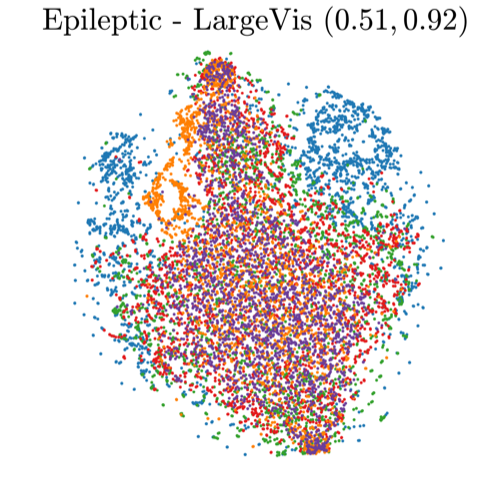}}
    \subfigure{\includegraphics[ width=0.31\textwidth]{./figs/epileptic_trimap.png}}
    \subfigure{\includegraphics[ width=0.31\textwidth]{./figs/epileptic_pca.png}}\hfill\\[-1mm]
    % \subfigure{\includegraphics[ width=0.31\textwidth]{./figs/news20_lv.png}}
    % \subfigure{\includegraphics[ width=0.31\textwidth]{./figs/news20_trimap.png}}
    % \subfigure{\includegraphics[ width=0.31\textwidth]{./figs/news20_pca.png}}\hfill\\[-1mm]
    % \subfigure{\includegraphics[ width=0.31\textwidth]{./figs/tabula_lv.png}}
    % \subfigure{\includegraphics[ width=0.31\textwidth]{./figs/tabula_trimap.png}}
    % \subfigure{\includegraphics[ width=0.31\textwidth]{./figs/tabula_pca.png}}\hfill\\
    \vspace{-0.3cm}
    \caption{Visualizations of different datasets using LargeVis, TriMap, and PCA. Each row corresponds to one dataset and each column represents one method. The values of nearest neighbor accuracy and global score are shown as a pair \texttt{(NN,GS)} on top of each figure.}
    \label{fig:results-lv}
    \vspace{-0.3cm}
\end{figure}

\setcounter{figure}{6}
\begin{figure}[t!]
\vspace{-1.6cm}
    \centering
    % \subfigure{\includegraphics[ width=0.31\textwidth]{./figs/coil20_lv.png}}
    % \subfigure{\includegraphics[ width=0.31\textwidth]{./figs/coil20_trimap.png}}
    % \subfigure{\includegraphics[ width=0.31\textwidth]{./figs/coil20_pca.png}}\hfill\\[-1mm]
    % \subfigure{\includegraphics[ width=0.31\textwidth]{./figs/usps_lv.png}}
    % \subfigure{\includegraphics[ width=0.31\textwidth]{./figs/usps_trimap.png}}
    % \subfigure{\includegraphics[ width=0.31\textwidth]{./figs/usps_pca.png}}\hfill\\[-1mm]
    % \subfigure{\includegraphics[ width=0.31\textwidth]{./figs/epileptic_lv.png}}
    % \subfigure{\includegraphics[ width=0.31\textwidth]{./figs/epileptic_trimap.png}}
    % \subfigure{\includegraphics[ width=0.31\textwidth]{./figs/epileptic_pca.png}}\hfill\\[-1mm]
    \subfigure{\includegraphics[ width=0.31\textwidth]{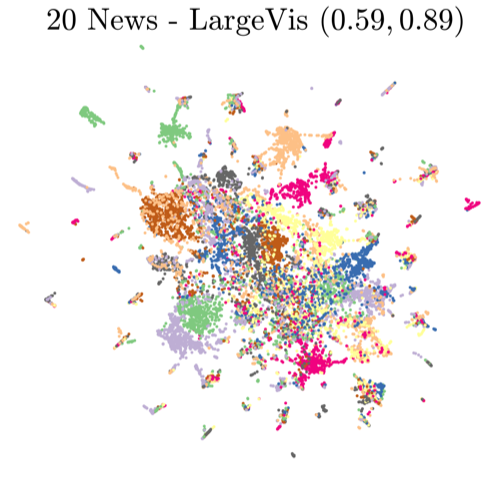}}
    \subfigure{\includegraphics[ width=0.31\textwidth]{./figs/news20_trimap.png}}
    \subfigure{\includegraphics[ width=0.31\textwidth]{./figs/news20_pca.png}}\hfill\\[-1mm]
    \subfigure{\includegraphics[ width=0.31\textwidth]{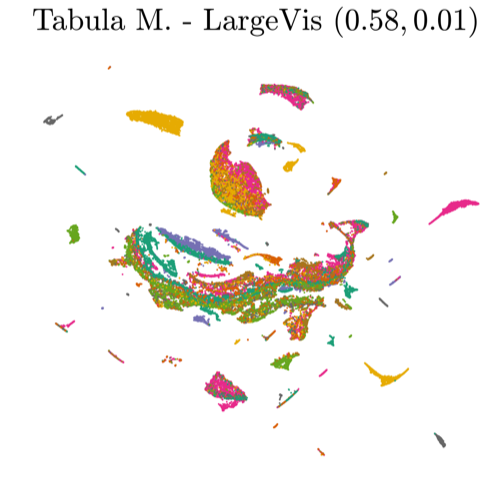}}
    \subfigure{\includegraphics[ width=0.31\textwidth]{./figs/tabula_trimap.png}}
    \subfigure{\includegraphics[ width=0.31\textwidth]{./figs/tabula_pca.png}}\hfill\\
    \subfigure{\includegraphics[ width=0.31\textwidth]{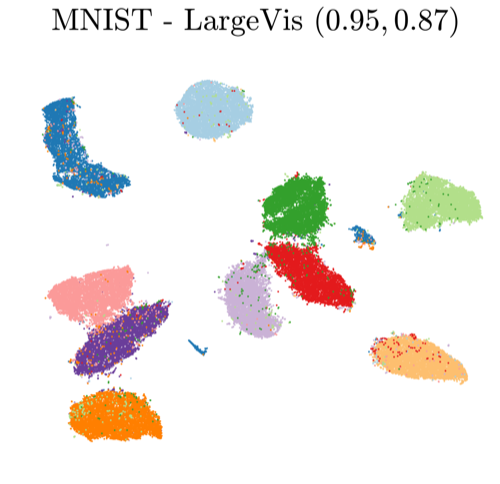}}
    \subfigure{\includegraphics[ width=0.31\textwidth]{./figs/mnist_trimap.png}}
    \subfigure{\includegraphics[ width=0.31\textwidth]{./figs/mnist_pca.png}}\hfill\\[-2mm]
    \vspace{-0.3cm}
    \caption{Visualizations of different datasets (continued) using LargeVis, TriMap, and PCA. Each row corresponds to one dataset and each column represents one method. The values of nearest neighbor accuracy and global score are shown as a pair \texttt{(NN,GS)} on top of each figure.}
    \label{fig:results-lv}
    \vspace{-0.3cm}
\end{figure}

\section{Discussion}
We briefly discuss the results of TriMap and draw a comparison to the other methods.

TriMap generally provides better global accuracy compared to the competing methods. It also successfully maintains the continuity of the underlying manifold. This can be seen from the COIL-20 result, where certain clusters are located farther away from the remaining clusters. However, the underlying structure for the main cluster resembles the one provided by the other methods. TriMap also preserves the continuous structure in the Fashion MNIST and the TV News datasets.

TriMap is also efficient in uncovering the possible outliers in the data. For instance, PCA reveals a large number of outliers in the Tabula Muris and the 360+K Lyrics datasets. These outliers are located far away from the main clusters in the TriMap results. However, the same points are located very close to the remaining points in the t-SNE results.

Additionally, both t-SNE and LargeVis tend to form spurious clusters by splitting the underlying connected manifold. This can be seen from the TV News results and the result of LargeVis on the Covertype dataset.

Finally, notice that in some cases GS fails to reflect the global accuracy of the embeddings. This can be seen from the low GS values for all methods on the Covertype dataset. GS may become uninformative when there exists a high degree of non-linearity in the data that cannot be reflected using PCA. GS also cannot reflect the accuracy of the embedding in uncovering single outliers.
Developing more accurate global measures for these scenarios is a future research direction.

\setcounter{figure}{6}
\begin{figure}
\vspace{-1.5cm}
    \centering
    \subfigure{\includegraphics[ width=0.31\textwidth]{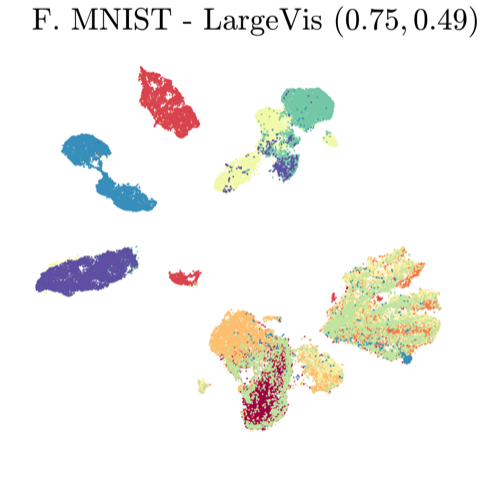}}
    \subfigure{\includegraphics[ width=0.31\textwidth]{./figs/fmnist_trimap.png}}
    \subfigure{\includegraphics[ width=0.31\textwidth]{./figs/fmnist_pca.png}}\hfill\\[-2mm]
    \subfigure{\includegraphics[ width=0.31\textwidth]{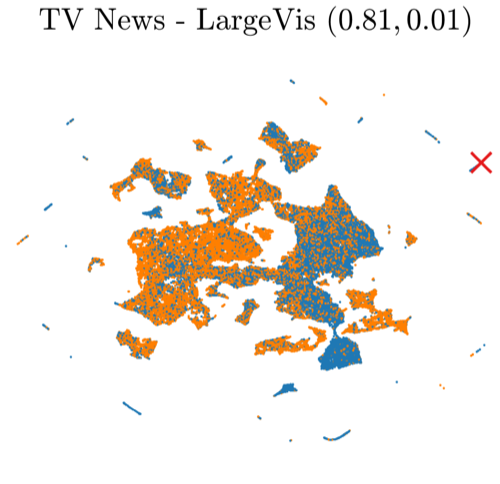}}
    \subfigure{\includegraphics[ width=0.31\textwidth]{./figs/tvnews_trimap.png}}
    \subfigure{\includegraphics[ width=0.31\textwidth]{./figs/tvnews_pca.png}}\hfill\\[-2mm]
    \subfigure{\includegraphics[ width=0.31\textwidth]{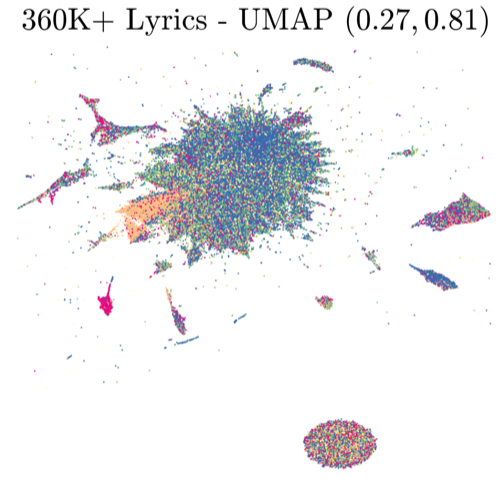}}
    \subfigure{\includegraphics[ width=0.31\textwidth]{./figs/lyrics_trimap.png}}
    \subfigure{\includegraphics[ width=0.31\textwidth]{./figs/lyrics_pca.png}}\hfill\\[-2mm]
    \vspace{-0.2cm}
    \caption{Visualizations of different datasets (continued) using LargeVis, TriMap, and PCA. Each row corresponds to one dataset and each column represents one method. The values of nearest neighbor accuracy and global score are shown as a tuple \texttt{(NN,GS)} on top of each figure.}
    % \vspace{-0.2cm}
\end{figure}

\begin{figure}[t!]
\vspace{-1.5cm}
    \centering
    \subfigure{\includegraphics[ width=0.32\textwidth]{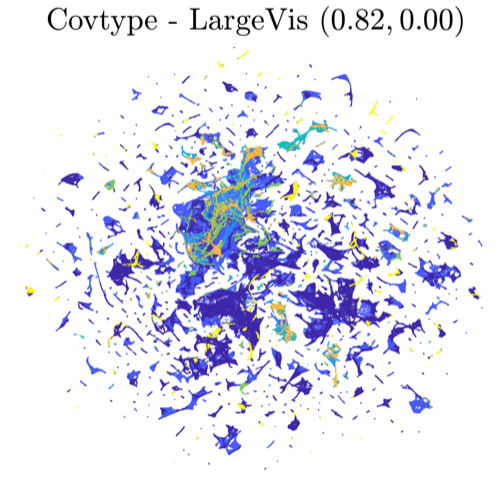}}
    \subfigure{\includegraphics[ width=0.32\textwidth]{./figs/covtype_trimap.png}}
    \subfigure{\includegraphics[ width=0.32\textwidth]{./figs/covtype_pca.png}}\hfill\\[-3mm]
    \subfigure{\includegraphics[ width=0.32\textwidth]{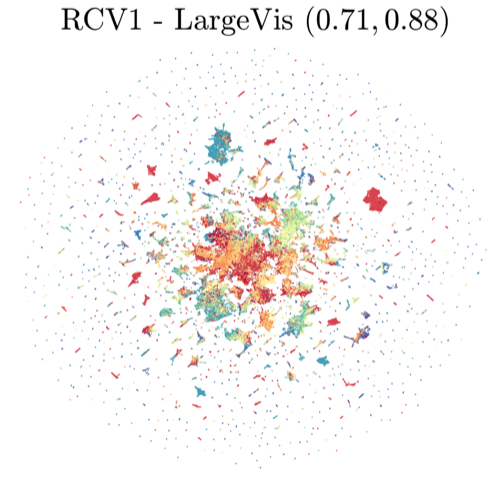}}
    \subfigure{\includegraphics[ width=0.32\textwidth]{./figs/rcv1_trimap.png}}
    \subfigure{\includegraphics[ width=0.32\textwidth]{./figs/rcv1_pca.png}}\hfill\\
    % \vspace{-0.2cm}
    \caption{Visualizations of Covertype and RCV1 datasets using LargeVis, TriMap, and PCA, and visualizations of the Character Font Images dataset using LargeVis, TriMap, and PCA. The values of nearest neighbor accuracy and global score are shown as a tuple \texttt{(NN,GS)} on top of each figure.}\label{fig:more-results-lv}
    % \vspace{0.3cm}
\end{figure}

\begin{figure}[b!]
% \vspace{-1.5cm}
    \centering
    \subfigure{\includegraphics[ width=0.35\textwidth]{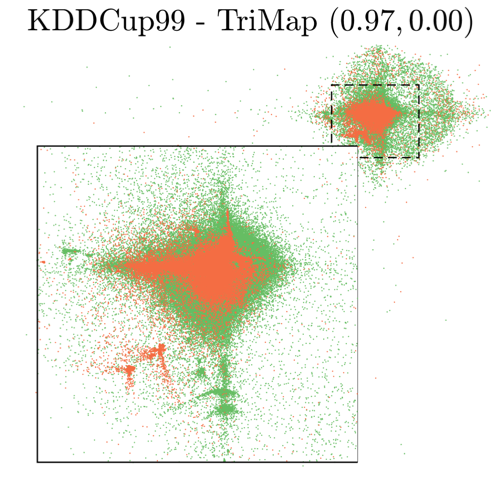}}
    \subfigure{\includegraphics[ width=0.35\textwidth]{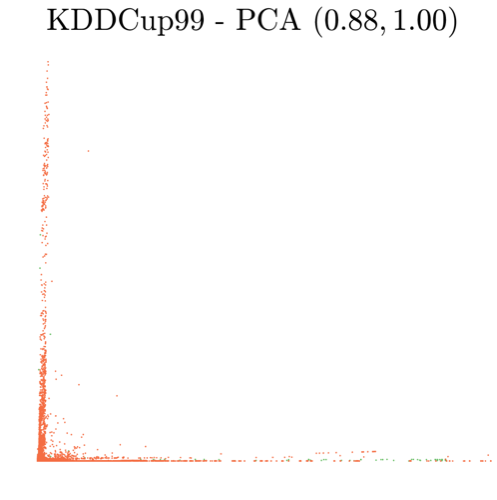}}\hfill\\
    \subfigure{\includegraphics[ width=0.35\textwidth]{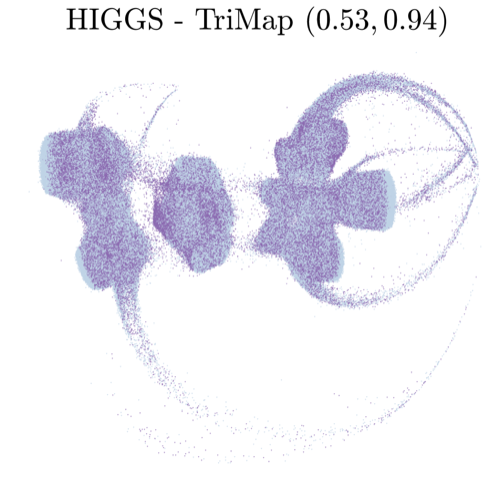}}
    \subfigure{\includegraphics[ width=0.35\textwidth]{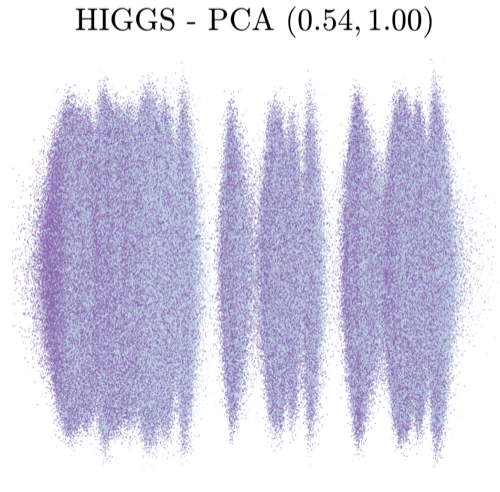}}\hfill\\
    \vspace{-0.2cm}
    \caption{Visualizations of KKDCup99 and HIGGS datasets TriMap and PCA. Each row corresponds to one dataset and each column represents one method. The values of nearest neighbor accuracy and global score are shown as a tuple \texttt{(NN,GS)} on top of each figure. TriMap shows more structure for both datasets than PCA. Note that GS is uninformative for the KDDCup99 dataset.}\label{fig:higgs}
    % \vspace{-0.2cm}
\end{figure}
\end{document}